%% file: PAPER.tex
\documentclass{article}

\usepackage{microtype}
\usepackage{graphicx}
\usepackage{subcaption}
\usepackage{booktabs} 
\usepackage{subcaption}

\usepackage{tikz}
\usetikzlibrary{backgrounds}
\usepackage{multirow}

\usepackage{hyperref}

\usepackage[preprint]{icml2026}

\usepackage{amsmath}
\usepackage{amssymb}
\usepackage{mathtools}
\usepackage{amsthm}
\usepackage{pgfplots}
\pgfplotsset{compat=1.18}  
\usepackage{acro}
\input{acronyms}

\usepackage[capitalize,noabbrev]{cleveref}

\theoremstyle{plain}

\theoremstyle{definition}

\theoremstyle{remark}

\usepackage[textsize=tiny]{todonotes}

\icmltitlerunning{Mechanistic Indicators for Steering}

\begin{document}

\twocolumn[
  \icmltitle{Mechanistic Indicators of Steering Effectiveness in Large Language Models}




  \icmlsetsymbol{equal}{*}

  \begin{icmlauthorlist}
    \icmlauthor{Mehdi Jafari}{cse,adms}
    \icmlauthor{Hao Xue}{hk,adms}
    \icmlauthor{Flora Salim}{cse,adms}
  \end{icmlauthorlist}

  \icmlaffiliation{cse}{School of Computer Science and Engineering, University of New South Wales, Sydney, Australia}
  \icmlaffiliation{adms}{ARC Centre of Excellence for Automated Decision-Making and Society, Melbourne, Australia}
  \icmlaffiliation{hk}{The Hong Kong University of Science and Technology, Guangzhou, China}

  \icmlcorrespondingauthor{Mehdi Jafari}{mehdi.jafari@unsw.edu.au}
  \icmlcorrespondingauthor{Flora Salim}{flora.salim@unsw.edu.au}

  \vskip 0.3in
]



\printAffiliationsAndNotice{}  
\newcommand{\gpt}{\texttt{ChatGPT-4o-mini}}
\newcommand{\gmn}{\texttt{Gemini-Flash-2.5}}


\begin{abstract}
Activation-based steering enables \acp{llm} to exhibit targeted behaviors by intervening on intermediate activations without retraining. Despite its widespread use, the mechanistic factors that govern when steering succeeds or fails remain poorly understood, as prior work has relied primarily on black-box outputs or \ac{llm}-based judges. In this study, we investigate whether the reliability of steering can be diagnosed using internal model signals. We focus on two information-theoretic measures: the entropy-derived \ac{nbf}, and the \ac{kl} divergence between steered activations and targeted concepts in the vocabulary space. We hypothesize that effective steering corresponds to structured entropy preservation and coherent \ac{kl} alignment across decoding steps. Building on a reliability study demonstrating high inter-judge agreement between two architecturally distinct \acp{llm}, we use \ac{llm}-generated annotations as ground truth and show that these mechanistic signals provide meaningful predictive power for identifying successful steering and estimating failure probability. We further introduce a stronger evaluation baseline for \ac{caa} and Sparse Autoencoder-based steering, the two most widely adopted activation-steering methods.\footnote{Code is available at \nolinkurl{https://github.com/cruiseresearchgroup/IntSteer}.}
\end{abstract}

\input{graphs/pipeline_styles}

\input{graphs/pipeline}
\input{sections/introduction}

\input{sections/formalization}

\input{sections/signals}

\input{graphs/KL_9b_Christian_rotation_act_5}
\input{sections/method}
\input{sections/experiments}

\input{sections/results}
\input{sections/related}

\input{sections/discussion}

\input{sections/limits}
\input{sections/conclusion}

\input{sections/impacts}
\input{sections/ack}
\nocite{langley00}
\newpage
\bibliography{example_paper}
\bibliographystyle{icml2026}

\newpage
\appendix
\onecolumn
\input{sections/app}

\end{document}

%% file: acronyms.tex
\DeclareAcronym{llm}{
  short = LLM,
  long  = large language model
}

\DeclareAcronym{nlp}{
  short = NLP,
  long  = natural language processing
}

\DeclareAcronym{caa}{
    short = CAA,
    long = Contrastive Activation Addition 
}

\DeclareAcronym{gpt}{
    short = ChatGPT-4o-mini,
    long = ChatGPT-4o-mini
}

\DeclareAcronym{gmn}{
    short = Gemini-Flash-2.5,
    long = Gemini-Flash-2.5
}

\DeclareAcronym{sae}{
    short = SAE,
    long = Sparse Autoencoder 
}
\DeclareAcronym{kl}{
  short = KL,
  long  = Kullback--Leibler
}

\DeclareAcronym{nbf}{
  short = NBF,
  long  = Normalized Branching Factor
}
\DeclareAcronym{icml}{
  short = ICML,
  long  = International Conference on Machine Learning
}

\DeclareAcronym{cot}{
  short = CoT,
  long  = chain-of-thought
}

%% file: graphs/pipeline_styles.tex
\tikzstyle{addition} = [
  circle,
  minimum size=0.5cm,
  text centered,
  fill=yellow!80!black,  
  draw=gray,
  line width=0.5mm
]

\tikzstyle{steer} = [
  circle,
  minimum size=0.5cm,
  text centered,
  fill=teal!15,
  draw=gray,
  line width=0.5mm
]

\tikzstyle{concept} = [rectangle, rounded corners, 
minimum width=2cm, 
minimum height=1cm,
text centered, 
fill=gray!30]

\tikzstyle{embed} = [rectangle, rounded corners, 
minimum width=2cm, 
minimum height=1cm,
text centered, 
draw=gray,
fill=gray!30]

\tikzstyle{unembed} = [rectangle, rounded corners, 
minimum width=2cm, 
minimum height=1cm,
text centered, 
draw=gray,
fill=gray!30]

\tikzstyle{logits} = [rectangle, rounded corners, 
minimum width=2cm, 
minimum height=1cm,
text centered, 
draw=gray,
fill=gray!30]

\tikzstyle{tokens} = [rectangle, rounded corners, 
minimum width=2cm, 
minimum height=1cm,
text centered, 
draw=gray,
fill=gray!30]

\tikzstyle{inout} = [rectangle, rounded corners, 
minimum width=2cm, 
minimum height=1cm,
text centered, 
fill=none,
draw=none]

\tikzstyle{max} = [rectangle, rounded corners, 
minimum width=3.75cm, 
minimum height=2.5cm,
text centered, 
line width=2pt, 
draw=green!60!black,  
dashed, 
fill=green!60!black,  
fill opacity=0.05]

\tikzstyle{kl} = [rectangle, rounded corners, 
minimum width=4.6cm, 
minimum height=5.1cm,
text centered, 
line width=2pt, 
draw=red!60!black,  
dashed, 
fill=red!60!black,  
fill opacity=0.05]

\tikzstyle{nbf} = [
  rectangle,
  rounded corners,
  minimum width=2.3cm,
  minimum height=1.3cm,
  line width=2pt,     
  text centered,
  draw=blue!60!black,      
  dashed,                  
  fill=blue!60!black,      
  fill opacity=0.05        
]

\tikzstyle{mlp} = [rectangle, rounded corners, 
minimum width=2.5cm, 
minimum height=1cm,
text centered, 
draw=gray,
fill=gray!30]

\tikzstyle{steer} = [rectangle, rounded corners, 
minimum width=2.5cm, 
minimum height=1cm,
text centered, 
fill=gray!30,
draw=gray,]

\tikzstyle{prompt} = [rectangle, rounded corners, 
minimum width=3cm, 
minimum height=1cm,
text centered, 
fill=None,
draw=None]

\tikzstyle{response} = [rectangle, rounded corners, 
minimum width=3cm, 
minimum height=1cm,
text centered, 
fill=None,
draw=None]

\tikzstyle{head} = [rectangle, rounded corners, 
minimum width=1cm, 
minimum height=1cm,
text centered, 
fill=gray!30,
draw=gray]

\tikzstyle{invis} = [
  draw=none,
  fill=black,
  text opacity=0,
  inner sep=0pt,
  outer sep=0pt
]

\tikzstyle{steering_block} = [rectangle, rounded corners, 
minimum width=4cm, 
minimum height=2cm,
text centered, 
fill=none,
draw=gray, 
line width=.5mm, 
]

\tikzstyle{transformer_block} = [rectangle, rounded corners, 
minimum width=5cm, 
minimum height=10cm,
text centered, 
fill=none,
draw=gray, 
line width=.5mm]

\tikzstyle{llm} = [rectangle, rounded corners, 
minimum width=3cm, 
minimum height=1cm,
text centered, 
fill=brown!30]

\tikzstyle{annotation} = [rectangle, rounded corners, 
minimum width=3cm, 
minimum height=1cm,
text centered, 
fill=brown!30]

\tikzstyle{regression} = [rectangle, rounded corners, 
minimum width=3cm, 
minimum height=1cm,
text centered, 
fill=brown!30]

\tikzstyle{bluebox} = [rectangle, rounded corners, 
minimum width=1cm, 
minimum height=5cm,
text centered, 
draw=blue!60!black,
dashed,
fill=blue!60!black, 
fill opacity=0.05]

\tikzstyle{greenbox} = [rectangle, rounded corners, 
minimum width=1cm, 
minimum height=5cm,
text centered, 
draw=green!60!black,
dashed,
fill=green!60!black, 
fill opacity=0.05]

\tikzstyle{redbox} = [rectangle, rounded corners, 
minimum width=1cm, 
minimum height=5cm,
text centered, 
draw=red!60!black,
dashed,
fill=red!60!black, 
fill opacity=0.05]

%% file: graphs/pipeline.tex
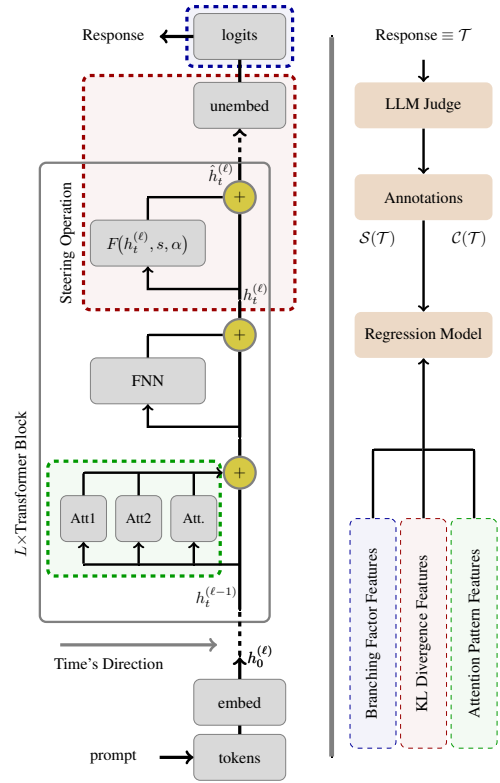
\begin{figure}[!ht]
  
  \vskip 0.2in
  \begin{center}\resizebox{0.8\columnwidth}{!}{%
    \resizebox{\columnwidth}{!}{%
    
\begin{tikzpicture}[node distance=2cm]

\begin{scope}[on background layer] 
    \node (bottom) [invis] {};
    \node (1) [invis, above of=bottom, yshift=-0.8cm] {};
    \node (2) [invis, above of=1] {};
    \node (max) [left of=2, max, xshift=-2.3cm, yshift=2cm] {};
    \node (3) [invis, above of=2, yshift=1cm] {};
    \node (bluebox) [bluebox, right of=2, xshift=-1.1cm, yshift=-.5cm, text opacity=1] {\rotatebox{90}{Branching Factor Features}}; 
    \node (redbox) [redbox, right of=2, yshift=-.5cm, text opacity=1] {\rotatebox{90}{KL Divergence Features}}; 
    \node (greenbox) [greenbox, right of=2, xshift=1.1cm, yshift=-.5cm, text opacity=1] {\rotatebox{90}{Attention Pattern Features}}; 
    \node (blueboxt) [above of=bluebox, yshift=2cm, invis] {}; 
    \node (redboxt) [above of=redbox, yshift=2cm, invis] {}; 
    \node (greenboxt) [above of=greenbox, yshift=2cm, invis] {}; 
    \node (4) [invis, above of=3, yshift=1cm] {};
    \node (5) [invis, above of=4, yshift=1cm] {};
    \node (kl) [left of=5, kl, xshift=-1.1cm, yshift=0.1cm] {};
    \node (6) [invis, above of=5] {};
    \node (llm) [llm, right of=6] {LLM Judge}; 
    \node (annotation) [annotation, right of=5] {Annotations}; 
    \node (regression) [regression, right of=4] {Regression Model}; 
    \node (7) [invis, above of=6, yshift=-0.5cm] {};
    \node (nbf) [left of=7, nbf] {};
    \node (tokens) [left of=bottom, tokens] {tokens};
    \node (prompt) [left of=tokens, inout, xshift=-0.75cm] {prompt};
    \node (embed) [left of=1, embed] {embed};
    \node (x0) [left of=1, invis, yshift=1cm] {};
    \node (x1) [left of=2, invis] {};
    \node (linkx1) [left of=2, invis, yshift=+1cm] {};
    \node (add1) [addition, left of=3] {$+$}; 
    \node (x2) [left of=3, invis, yshift=.75cm] {};
    \node (linkx2) [left of=3, invis, yshift=+1cm] {};
    \node (add2) [addition, left of=4] {$+$}; 
    \node (x3) [left of=4, invis, yshift=.75cm] {};
    \node (linkx3) [left of=4, invis, yshift=+1cm] {};
    \node (add3) [addition, left of=5] {$+$}; 
    \node (x4) [left of=5, invis, yshift=.75cm] {};
    \node (unembed) [left of=6, unembed] {unembed};
    \node (logits) [left of=7, logits] {logits};
    \node (response) [left of=logits, xshift=-.75cm, inout] {Response};
    \node (response2) [right of=7, inout] {{Response $\equiv \mathcal{T}$}};
    
    \node (head1) [left of=3, head, yshift=-1cm, xshift=-1cm] {\small Att.}; 
    \node (head2) [left of=3, head, yshift=-1cm, xshift=-2.2cm] {\small Att2}; 
    \node (head3) [left of=3, head, yshift=-1cm, xshift=-3.4cm] {\small Att1}; 
    
    \node (head1b) [left of=3, invis, yshift=-2cm, xshift=-1cm] {}; 
    \node (head2b) [left of=3, invis, yshift=-2cm, xshift=-2.2cm] {}; 
    \node (head3b) [left of=3, invis, yshift=-2cm, xshift=-3.4cm] {}; 

    \node (head1t) [left of=3, invis, xshift=-1cm] {}; 
    \node (head2t) [left of=3, invis, xshift=-2.2cm] {}; 
    \node (head3t) [left of=3, invis, xshift=-3.4cm] {}; 

    \node (mlp) [left of=4, mlp, yshift=-1cm, xshift=-2cm] {FNN}; 
    \node (mlpb) [left of=4, invis, yshift=-2cm, xshift=-2cm] {}; 
    \node (mlpt) [left of=4, invis, xshift=-2cm] {}; 

    \node (steer) [left of=5, steer, yshift=-1cm, xshift=-2cm] {$F\!\big(h^{(\ell)}_t, s, \alpha\big)$}; 
    \node (steerb) [left of=5, invis, yshift=-2cm, xshift=-2cm] {}; 
    \node (steert) [left of=5, invis, xshift=-2cm] {}; 

    \node (tb) [left of=x2, transformer_block, yshift=1cm, xshift=0.15cm] {}; 

    \node [rotate=90, anchor=south, left of=tb,yshift=2.9cm] {$L \times $Transformer Block};
    \node [anchor=south, above of=add2, xshift=+.35cm, yshift=-1.1cm] {$h^{(\ell)}_t$};
    \node [anchor=south, above of=embed, xshift=-.5cm, yshift=+.3cm] {$h^{(\ell-1)}_t$};
    \node [anchor=south, above of=embed, xshift=+.45cm, yshift=-1cm] {$h^{(\ell)}_0$};
    \node [anchor=south, above of=embed, xshift=+.45cm, yshift=-1cm] {$h^{(\ell)}_0$};
    \node [anchor=south, above of=add3, xshift=-.4cm, yshift=-1.5cm] {$\hat{h}^{(\ell)}_t$};
    \node [anchor=south, below of=annotation, xshift = -1cm, yshift=1.1cm] {$\mathcal{S}(\mathcal{T})$};
    \node [anchor=south, below of=annotation, xshift = 1cm, yshift=1.1cm] {$\mathcal{C}(\mathcal{T})$};
    \node [rotate=90, anchor=south, left of=steer, xshift=2cm, yshift=1.75cm] {Steering Operation};
    \node [anchor=south, below of=tb, xshift=-1cm, yshift=-3.9cm] {Time's Direction};
    
\end{scope}
\draw [->, line width=2pt] (prompt) -- (tokens); 
\draw [-, line width=2pt] (tokens) -- (embed); 
\draw [->, line width=2pt] (embed) -- (x0); 
\draw [-, line width=2pt, dashed] (x0) -- (x1); 
\draw [-, line width=2pt] (x1) -- (add1); 
\draw [-, line width=2pt] (add1) -- (x2); 
\draw [-, line width=2pt] (x2) -- (add2); 
\draw [-, line width=2pt] (add2) -- (x3); 
\draw [-, line width=2pt] (x3) -- (add3); 
\draw [-, line width=2pt] (add3) -- (x4); 
\draw [->, line width=2pt, dashed] (x4) -- (unembed); 
\draw [-, line width=2pt] (unembed) -- (logits); 
\draw [->, line width=2pt] (logits) -- (response); 

\draw [->, line width=1.5pt] (head1b) -- (head1.south); 
\draw [->, line width=1.5pt] (head2b) -- (head2.south); 
\draw [->, line width=1.5pt] (head3b) -- (head3.south); 
\draw [-, line width=1.5pt] (linkx1) -- (head3b); 

\draw [-, line width=1.5pt] (head1.north) -- (head1t); 
\draw [-, line width=1.5pt] (head2.north) -- (head2t); 
\draw [-, line width=1.5pt] (head3.north) -- (head3t); 
\draw [->, line width=1.5pt] (head3t) -- (add1); 

\draw [->, line width=1.5pt] (linkx2) -| (mlp.south); 
\draw [-, line width=1.5pt] (mlp.north) -- (mlpt); 
\draw [-, line width=1.5pt] (mlpt) -- (add2); 

\draw [->, line width=1.5pt] (linkx3) -| (steer.south); 
\draw [-, line width=1.5pt] (steer.north) -- (steert); 
\draw [-, line width=1.5pt] (steert) -- (add3); 

\draw [->, line width=1.5pt] (response2) -- (llm); 
\draw [->, line width=1.5pt] (llm) -- (annotation); 
\draw [->, line width=1.5pt] (annotation) -- (regression); 

\draw [-, line width=1.5pt] (bluebox) -- (blueboxt); 
\draw [-, line width=1.5pt] (greenbox) -- (greenboxt); 
\draw [-, line width=1.5pt] (redbox) -- (redboxt);
\draw [-, line width=1.5pt] (redboxt) -- (blueboxt); 
\draw [-, line width=1.5pt] (redboxt) -- (greenboxt);
\draw [->, line width=1.5pt] (redboxt) -- (regression);

\draw [-, line width=3pt, draw=gray] (bottom) -- (7);

\draw [->, line width=2pt, draw=gray, transform canvas={yshift=-2.75cm}] (head3.west) -- (head1.east);

\end{tikzpicture}
}
}
  \end{center}
  \caption{Overall pipeline of the proposed method. (Left) An abstract schematic of the \ac{llm}, illustrating the extraction of three distinct feature sets (blue, red, and green blocks). (Right) An overview of how the extracted features are utilized within the regression framework. }
  \label{fig:pipeline}

\end{figure}

%% file: sections/introduction.tex
\section{Introduction}

Recent work \cite{turnerSteeringLanguageModels2024a, nguyenMultiAttributeSteeringLanguage, wangImprovingLLMReasoning2025} shows that \acp{llm} can be steered via intervention vectors in the residual stream, enabling controllable generation without retraining. Major approaches include \ac{caa} \cite{rimsky-etal-2024-steering, haoPatternsMechanismsContrastive2025} and \acp{sae} \cite{lieberumGemmaScopeOpen2024, chalnevImprovingSteeringVectors2024, joshiIdentifiableSteeringSparse2025, choCorrSteerSteeringImproves2025}, which bias outputs by enriching residual representations. Despite their growing use \cite{sooInterpretableSteeringLarge2025, aradSAEsAreGood2025, wangEnhancingLLMSteering2025, sunHyperDASAutomatingMechanistic2025}, the mechanistic factors behind steering success remain unclear, leaving it as a largely heuristic control method rather than a principled intervention.


Although quantitative metrics can assess steering \cite{wangImprovingLLMReasoning2025}, most tasks involve free-text generation, where evaluation practices limit our understanding of reliability. Many studies \cite{wuAxBenchSteeringLLMs2025, chalnevImprovingSteeringVectors2024, sooInterpretableSteeringLarge2025, thakurJudgingJudgesEvaluating2025, liGenerationJudgmentOpportunities2025} use \acp{llm} as automated judges, reducing human annotation but outsourcing reliability to another opaque model. This raises the question of whether such evaluations reflect true steering success or artifacts of the judge’s biases.


Mechanistic interpretability research \cite{raiPracticalReviewMechanistic2025, sunHyperDASAutomatingMechanistic2025, bereskaMechanisticInterpretabilityAI2024, sharkeyOpenProblemsMechanistic2025} provides tools to analyze neural network internals, effective in tasks like knowledge conflict detection \cite{zhaoAnalysingResidualStream2024}, theory-of-mind monitoring \cite{jafariEnhancingConversationalAgents2025}, world modeling \cite{karvonenEmergentWorldModels2024, gurneeLanguageModelsRepresent2024}, and user modeling \cite{chenDesigningDashboardTransparency2024}. Yet, these tools are underused for steering evaluation, which remains outcome-focused. For instance internal signals, such as disrupted attention patterns, can indicate a high risk of failure by revealing when the model loses track of long-term dependencies, and these signals are directly extractable during generation.


In this work, we adopt an empirical, interpretability-driven perspective and argue that steering effectiveness or failure, can be inferred directly from an \ac{llm}'s internal mechanistic signals. We frame steering as a natural component of the \ac{llm}, analogous to attention heads and feed-forward multilayer perceptrons, which read information from the residual stream and write enriched representations back into it. Under this view, the performance of a steering block can be analyzed using existing mechanistic interpretability toolsets, supported by established linguistic and interpretability insights into \ac{llm} behavior.


We focus on two information-theoretic signals: \ac{nbf}, based on vocabulary entropy, and \ac{kl} proximity between behavior-specific vocabularies and steered versus unsteered outputs (Section \ref{signals}). These serve as mechanistic indicators of how steering reshapes model behavior across layers and decoding steps. We hypothesize that effective steering preserves structured entropy while aligning \ac{kl} with steering vectors. To test this, we evaluate whether these signals can predict steering quality or failure during generation without external evaluators.


Due to the infeasibility of large-scale human annotation and the inherently ambiguous nature of qualitative text assessment for this task, we treat \ac{llm}-generated annotations as ground truth. To mitigate concerns regarding evaluator bias and annotation reliability, we conduct a dedicated reliability study prior to our empirical and interpretability analyses (Section~\ref{reliability}). This study employs two comparably capable but architecturally distinct \ac{llm} judges—\gpt \space and \gmn—developed by different organizations. The results demonstrate consistent inter-judge agreement across a broad set of experimental conditions (different models, targeted behaviors, and steering methods), supporting the use of \acp{llm} as evaluators in our subsequent analyses.

Our contributions are as follows:

\begin{enumerate}

\item We empirically assess the reliability of \acp{llm} judges as qualitative annotators for steering evaluation across diverse experimental settings, including different target models, steering vectors, and steering functions (2{,}304 experiments).

\item We provide qualitative analysis and a mechanistic account of activation-based steering by examining \ac{nbf}, derived from entropy, together with \ac{kl} divergence dynamics across layers and decoding steps (cherry-picked examples are provided).

\item We demonstrate that steering quality scores can be predicted from internal model signals with reasonable accuracy.

\item Based on qualitative analysis, we introduce a stronger benchmarking baseline for activation-based steering methods, focusing on \ac{caa} and \ac{sae}-based steering as the most widely adopted approaches.

\end{enumerate}

\input{graphs/london_add}
\input{graphs/london_rot}


The paper is organized as follows. Section~\ref{formalization} introduces steering preliminaries and notation. Section~\ref{signals} presents mechanistic signals for analyzing steering. Section~\ref{method} describes our methodology, followed by the experimental setup (Section~\ref{experiments}) and results (Section~\ref{results}). Related work is in Section~\ref{related}, and Sections~\ref{discussion}, \ref{limits}, and \ref{conclude} provide discussion, limitations, and conclusions, respectively.

%% file: graphs/london_add.tex
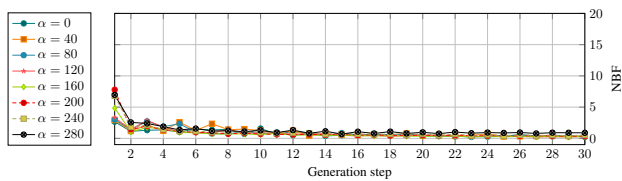
\begin{figure}[ht]
  \vskip 0.2in
  \begin{center}
    \resizebox{\columnwidth}{!}{%
    \begin{tikzpicture}
    \begin{axis}[
        width=14cm,
        height=5.05cm,
        grid=both,
        xmin=1,
        xmax=30,
        ymin=-1,
        ymax=20,
        xlabel={Generation step},
        ylabel={\ac{nbf}},
        legend style={
            at={(-0.05,0.5)},
            anchor=east
        },
        yticklabel pos=right,
        ylabel near ticks,
        legend columns=1,
        legend cell align=left,
        cycle list name=exotic
    ]

    \addplot coordinates {
    (1, 2.66) (2, 1.12) (3, 1.31) (4, 1.35) (5, 1.04) (6, 1.49) (7, 1.36) (8, 1.43) (9, 0.7) (10, 1.58) (11, 0.65) (12, 1.13) (13, 0.8) (14, 0.39) (15, 0.5) (16, 0.51) (17, 0.62) (18, 0.45) (19, 0.65) (20, 0.41) (21, 0.36) (22, 0.32) (23, 0.27) (24, 0.27) (25, 0.32) (26, 0.6) (27, 0.4) (28, 0.46) (29, 0.31) (30, 0.25)
    };
    \addlegendentry{$\alpha=0$}
    
    \addplot coordinates {
    (1, 2.93) (2, 1.18) (3, 2.16) (4, 1.21) (5, 2.6) (6, 1.25) (7, 2.34) (8, 1.4) (9, 1.48) (10, 1.26) (11, 0.83) (12, 1.24) (13, 0.45) (14, 0.69) (15, 0.59) (16, 0.55) (17, 0.67) (18, 0.58) (19, 0.78) (20, 0.44) (21, 0.57) (22, 0.38) (23, 0.46) (24, 0.73) (25, 0.3) (26, 0.32) (27, 0.37) (28, 0.41) (29, 0.38) (30, 0.39)
    
    };
    \addlegendentry{$\alpha=40$}
    
    \addplot coordinates {
    (1, 3.02) (2, 1.4) (3, 2.73) (4, 1.89) (5, 2.31) (6, 1.01) (7, 0.75) (8, 0.72) (9, 1.1) (10, 0.76) (11, 0.95) (12, 0.48) (13, 0.75) (14, 0.44) (15, 0.85) (16, 0.58) (17, 0.58) (18, 0.61) (19, 0.66) (20, 0.47) (21, 0.49) (22, 0.35) (23, 0.23) (24, 0.4) (25, 0.46) (26, 0.49) (27, 0.31) (28, 0.32) (29, 0.42) (30, 0.27)
    
    };
    \addlegendentry{$\alpha=80$}
    
    \addplot coordinates {
    (1, 3.29) (2, 1.44) (3, 2.88) (4, 1.81) (5, 1.03) (6, 0.9) (7, 0.87) (8, 1.41) (9, 0.85) (10, 0.85) (11, 0.84) (12, 0.68) (13, 0.72) (14, 0.75) (15, 0.49) (16, 0.39) (17, 0.48) (18, 0.33) (19, 0.44) (20, 0.48) (21, 0.45) (22, 0.31) (23, 0.28) (24, 0.38) (25, 0.29) (26, 0.29) (27, 0.24) (28, 0.25) (29, 0.25) (30, 0.3)
    
    };
    \addlegendentry{$\alpha=120$}
    
    \addplot coordinates {
    (1, 4.85) (2, 1.06) (3, 1.73) (4, 1.54) (5, 0.94) (6, 0.89) (7, 0.71) (8, 0.7) (9, 0.71) (10, 0.61) (11, 0.8) (12, 0.5) (13, 0.56) (14, 0.47) (15, 0.5) (16, 0.45) (17, 0.39) (18, 0.43) (19, 0.36) (20, 0.34) (21, 0.35) (22, 0.36) (23, 0.26) (24, 0.28) (25, 0.27) (26, 0.31) (27, 0.25) (28, 0.32) (29, 0.22) (30, 0.34)
    
    };
    \addlegendentry{$\alpha=160$}
    
    \addplot coordinates {
    (1, 7.79) (2, 1.44) (3, 1.98) (4, 1.35) (5, 1.16) (6, 0.98) (7, 0.9) (8, 0.74) (9, 0.85) (10, 0.69) (11, 0.64) (12, 0.61) (13, 0.64) (14, 0.59) (15, 0.66) (16, 0.52) (17, 0.56) (18, 0.46) (19, 0.5) (20, 0.51) (21, 0.45) (22, 0.51) (23, 0.43) (24, 0.44) (25, 0.43) (26, 0.4) (27, 0.44) (28, 0.38) (29, 0.36) (30, 0.34)
    
    };
    \addlegendentry{$\alpha=200$}
    
    \addplot coordinates {
    (1, 6.8) (2, 1.82) (3, 2.13) (4, 1.44) (5, 1.18) (6, 1.16) (7, 1.0) (8, 1.02) (9, 1.01) (10, 0.96) (11, 0.83) (12, 0.92) (13, 0.84) (14, 0.73) (15, 0.68) (16, 0.72) (17, 0.64) (18, 0.7) (19, 0.58) (20, 0.73) (21, 0.48) (22, 0.7) (23, 0.45) (24, 0.62) (25, 0.41) (26, 0.57) (27, 0.44) (28, 0.54) (29, 0.41) (30, 0.53)
    
    };
    \addlegendentry{$\alpha=240$}
    
    \addplot coordinates {
    (1, 6.96) (2, 2.56) (3, 2.39) (4, 1.87) (5, 1.35) (6, 1.57) (7, 1.18) (8, 1.23) (9, 1.08) (10, 1.28) (11, 0.92) (12, 1.32) (13, 0.82) (14, 1.14) (15, 0.69) (16, 1.07) (17, 0.79) (18, 1.08) (19, 0.79) (20, 0.97) (21, 0.76) (22, 1.01) (23, 0.83) (24, 0.93) (25, 0.87) (26, 0.94) (27, 0.79) (28, 0.91) (29, 0.9) (30, 0.89)

    };
    \addlegendentry{$\alpha=280$}
    
    \end{axis}
    \end{tikzpicture}
    }
  \end{center}
  \caption{unsuccessful steering example. The best performance is achieved $0.06$, corresponding to the Gemma 2–2B model for the London concept using the addition steering function with the \ac{sae} extraction method. No clear increase in \ac{nbf} is observed as the steering intensity $\alpha$ increases.}
  \label{fig:bad_nbf}

\end{figure}

%% file: graphs/london_rot.tex
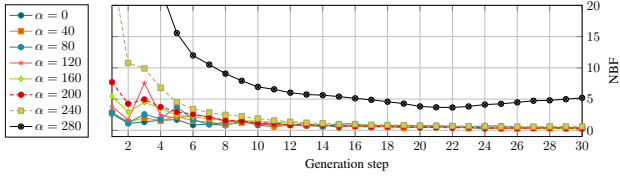
\begin{figure}[ht]
  \vskip 0.2in
  \begin{center}
    \resizebox{\columnwidth}{!}{%
    \begin{tikzpicture}
    \begin{axis}[
        width=14cm,
        height=5.05cm,
        grid=both,
        xmin=1,
        xmax=30,
        ymin=-1,
        ymax=20,
        xlabel={Generation step},
        ylabel={\ac{nbf}},
        legend style={
            at={(-0.05,0.5)},
            anchor=east
        },
        yticklabel pos=right,
        ylabel near ticks,
        legend columns=1,
        legend cell align=left,
        cycle list name=exotic
    ]

    \addplot coordinates {
    (1, 2.66) (2, 1.1) (3, 1.34) (4, 1.62) (5, 1.7) (6, 0.84) (7, 1.01) (8, 0.78) (9, 1.49) (10, 0.82) (11, 0.58) (12, 0.86) (13, 0.71) (14, 0.91) (15, 0.46) (16, 0.86) (17, 0.83) (18, 0.7) (19, 0.34) (20, 0.6) (21, 0.75) (22, 0.59) (23, 0.45) (24, 0.27) (25, 0.4) (26, 0.32) (27, 0.54) (28, 0.4) (29, 0.3) (30, 0.23)
    
    };
    \addlegendentry{$\alpha=0$}
    
    \addplot coordinates {
    (1, 2.79) (2, 1.3) (3, 2.03) (4, 1.54) (5, 2.26) (6, 1.53) (7, 1.31) (8, 0.92) (9, 1.21) (10, 1.2) (11, 0.71) (12, 0.82) (13, 0.81) (14, 0.72) (15, 0.76) (16, 0.66) (17, 0.57) (18, 0.56) (19, 0.39) (20, 0.65) (21, 0.52) (22, 0.65) (23, 0.33) (24, 0.52) (25, 0.3) (26, 0.28) (27, 0.37) (28, 0.34) (29, 0.34) (30, 0.42)
    
    };
    \addlegendentry{$\alpha=40$}
    
    \addplot coordinates {
    (1, 2.85) (2, 1.23) (3, 2.57) (4, 1.81) (5, 3.77) (6, 1.72) (7, 0.93) (8, 1.28) (9, 1.48) (10, 1.08) (11, 1.14) (12, 0.85) (13, 0.81) (14, 0.59) (15, 1.03) (16, 1.04) (17, 0.5) (18, 0.77) (19, 0.81) (20, 0.76) (21, 0.67) (22, 0.6) (23, 0.49) (24, 0.72) (25, 0.69) (26, 0.52) (27, 0.43) (28, 0.44) (29, 0.33) (30, 0.46)
    
    };
    \addlegendentry{$\alpha=80$}
    
    \addplot coordinates {
    (1, 3.81) (2, 1.69) (3, 7.52) (4, 2.45) (5, 1.89) (6, 2.48) (7, 2.19) (8, 1.4) (9, 1.48) (10, 0.94) (11, 0.94) (12, 0.86) (13, 0.73) (14, 0.74) (15, 0.84) (16, 0.68) (17, 0.59) (18, 0.75) (19, 0.48) (20, 0.52) (21, 0.54) (22, 0.53) (23, 0.46) (24, 0.41) (25, 0.34) (26, 0.32) (27, 0.3) (28, 0.36) (29, 0.39) (30, 0.28)
    
    };
    \addlegendentry{$\alpha=120$}
    
    \addplot coordinates {
    (1, 5.46) (2, 2.95) (3, 4.47) (4, 3.54) (5, 2.39) (6, 2.17) (7, 1.81) (8, 1.58) (9, 1.5) (10, 1.52) (11, 1.27) (12, 1.29) (13, 1.02) (14, 0.83) (15, 0.79) (16, 0.68) (17, 0.66) (18, 0.65) (19, 0.62) (20, 0.47) (21, 0.49) (22, 0.49) (23, 0.49) (24, 0.4) (25, 0.38) (26, 0.35) (27, 0.43) (28, 0.31) (29, 0.28) (30, 0.35)
    
    };
    \addlegendentry{$\alpha=160$}
    
    \addplot coordinates {
    (1, 7.71) (2, 4.25) (3, 4.94) (4, 3.73) (5, 2.95) (6, 2.59) (7, 2.09) (8, 1.6) (9, 1.51) (10, 1.2) (11, 1.13) (12, 0.88) (13, 0.91) (14, 0.79) (15, 0.72) (16, 0.66) (17, 0.63) (18, 0.57) (19, 0.59) (20, 0.45) (21, 0.47) (22, 0.39) (23, 0.45) (24, 0.39) (25, 0.37) (26, 0.37) (27, 0.36) (28, 0.36) (29, 0.37) (30, 0.3)
    
    };
    \addlegendentry{$\alpha=200$}
    
    \addplot coordinates {
    (1, 25.98) (2, 10.77) (3, 9.9) (4, 6.82) (5, 4.51) (6, 3.41) (7, 2.85) (8, 2.48) (9, 2.25) (10, 1.89) (11, 1.58) (12, 1.38) (13, 1.22) (14, 1.12) (15, 0.98) (16, 0.96) (17, 0.89) (18, 0.85) (19, 0.82) (20, 0.74) (21, 0.72) (22, 0.69) (23, 0.68) (24, 0.64) (25, 0.6) (26, 0.61) (27, 0.59) (28, 0.62) (29, 0.62) (30, 0.62)
    
    };
    \addlegendentry{$\alpha=240$}
    
    \addplot coordinates {
    (1, 54.85) (2, 53.66) (3, 34.77) (4, 22.69) (5, 15.56) (6, 11.99) (7, 10.52) (8, 9.07) (9, 7.93) (10, 6.95) (11, 6.56) (12, 6.02) (13, 5.73) (14, 5.62) (15, 5.4) (16, 5.12) (17, 4.88) (18, 4.57) (19, 4.27) (20, 3.82) (21, 3.68) (22, 3.65) (23, 3.82) (24, 4.12) (25, 4.24) (26, 4.47) (27, 4.73) (28, 4.8) (29, 4.98) (30, 5.2)
    
    };
    \addlegendentry{$\alpha=280$}
    
    \end{axis}
    \end{tikzpicture}
    
    }
  \end{center}
  \caption{Successful steering example. The best performance is achieved $0.24$, corresponding to the Gemma 2–2B model for the London concept using the rotational steering function with the \ac{sae} extraction method. A clear increase in \ac{nbf} is observed as the steering intensity $\alpha$ increases.}
  \label{fig:good_nbf}
\end{figure}

%% file: sections/formalization.tex
\section{Problem Setup and Definitions} 
\label{formalization}

\subsection{Model and Notation}

Let $M_{\theta}$ denote an autoregressive language model with parameters $\theta$. Given an input prompt $X = (x_1,\dots,x_n)$ and a previously generated token sequence $y_{<t} = (y_1,\dots,y_{t-1})$, the model defines a next-token distribution
\[
P_{\theta}(y_t \mid X, y_{<t}) \;=\; \mathrm{softmax}(z_t),
\]
where $z_t \in \mathbb{R}^{|V|}$ denotes the logits at generation step $t$, and $V$ denotes the vocabulary. Figure~\ref{fig:pipeline} (left) provides a high-level visualization of such a model.

For analysis, attention is restricted to an \emph{effective vocabulary} $V_{\mathrm{eff}}(t) \subseteq V$, defined as the set of the $N$ most probable tokens under $P_{\theta}(\cdot \mid X, y_{<t})$. 
Let $d$ denote the residual stream dimension. The final-layer hidden state at step $t$ is denoted by $h^{(L)}_t \in \mathbb{R}^d$. The unembedding block of the language model is represented as an affine transformation

\begin{equation} \label{eq:unembed}
z_t \;=\; U\,h^{(L)}_t + b_U,
\end{equation}

where $U \in \mathbb{R}^{|V| \times d}$ and $b_U \in \mathbb{R}^{|V|}$.

\paragraph{Transformer architecture.}
The model consists of a stack of $L$ transformer layers acting on the residual stream. Let $h^{(\ell)}_t \in \mathbb{R}^d$ denote the residual representation at layer $\ell$ and step $t$, with $h^{(0)}_t$ the input embedding (possibly including position encodings). Each transformer layer is composed of multi-head self-attention (MHA) and a position-wise feed-forward network (FFN), together with residual connections and normalization (pre- or post-layer normalization depending on the variant).

The scaled dot-product attention is

\begin{equation}
\label{eq:attention}
\mathrm{Attn}(Q,K,V) \;=\; \mathrm{softmax}\!\left(\frac{QK^{\top}}{\sqrt{d_k}}\right)V,
\end{equation}

where $Q = H W_Q$, $K = H W_K$, $V = H W_V$ for a sequence representation $H$ and learned projections $W_Q, W_K, W_V$; $d_k$ is the key dimension. The FFN acts position-wise as
\[
\mathrm{FFN}(h) \;=\; \sigma\!\big(h W_1 + b_1\big)\, W_2 + b_2,
\]
where $\sigma(\cdot)$ denotes the GELU nonlinearity, and $W_1, W_2, b_1, b_2$ are learned parameters. The update on residual stream is 
\[
h^{(\ell)}_t \;\leftarrow\; h^{(\ell-1)}_t + \Delta^{(\ell-1)}_t,
\]
where $\Delta^{(\ell)}_t$ is the contribution from the combination of MHA and FFN (and normalization applied per architecture) at layer $l$.

\subsection{Steering as Residual Intervention}

We formalize steering as an intervention on the residual stream, analogous to standard layer operations. Let $c$ denote a control signal encoding a desired attribute or behavior, and let $g(.)$ be a (learned or predefined) mapping from control signals to the residual space,
\[
s \;=\; g(c) \in \mathbb{R}^d,
\]
where $d$ is the residual stream dimension. Let $\mathcal{L} \subseteq \{1,\dots,L\}$ be the subset of layers at which steering is applied, and let $\alpha \in \mathbb{R}$ be a scalar controlling intervention strength. For time step $t$ and any $\ell \in \mathcal{L}$, we define the residual steering update
\begin{equation}
\label{eq:steering_fnc}
\tilde{h}^{(\ell)}_t \;\leftarrow\; F\!\big(h^{(\ell)}_t, s, \alpha\big),
\end{equation}

where $F(\cdot)$ is an intervention function that adds the original residual representation $h^{(\ell)}_t$ with its contribution $\hat{\Delta}^{(\ell)}_t$. 

\[
\hat{\Delta}^{(\ell)}_t = F\!\big(h^{(\ell)}_t, s, \alpha\big) - h^{(\ell)}_t
\]
\input{graphs/heatmap_2b}
\subsection{Coherence and Steering Objectives}
Steering a generative model introduces a fundamental trade-off between preserving linguistic coherence and enforcing alignment with a target attribute encoded by the steering vector $s$. this researrch formalize this trade-off using two scalar evaluation functions computed on the fully generated text $\mathcal{T}$. Let $\mathcal{S}(\mathcal{T}) \in [0,1]$ denote a \emph{steering score}, which quantifies the extent to which the output reflects the desired attribute represented by $s$. Let $\mathcal{C}(\mathcal{T}) \in [0,1]$ denote a \emph{coherence score}, which serves as a proxy for fluency and semantic consistency. These objectives are typically in tension: increasing the steering strength $\alpha$ may enhance attribute expression while degrading coherence. Consequently, steering can be framed as an optimization problem under competing criteria, balancing behavioral control against the preservation of the model’s generative competence.

%% file: graphs/heatmap_2b.tex
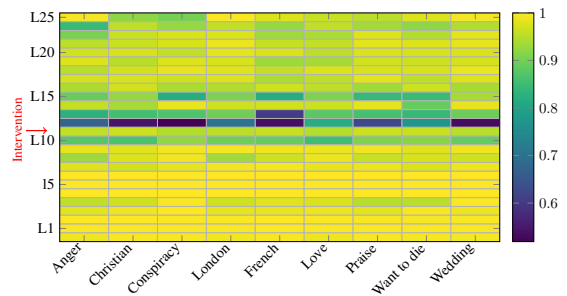
\begin{figure}[ht]
  \vskip 0.2in
  \begin{center}
    \resizebox{0.9\columnwidth}{!}{%
    \begin{tikzpicture}
      \begin{axis}[
        width=12cm,               
        height=7cm,               
        enlargelimits=false,
        colorbar,
        colormap/viridis, 
        xtick={0, 1, 2, 3, 4, 5, 6, 7, 8}, 
        x tick label style={rotate=45, anchor=east}, 
        xticklabels={Anger, Christian, Conspiracy, London, French, Love, Praise, Want to die, Wedding}, 
        ytick={0, 4, 9, 14, 19, 24},
        yticklabels={L25, L20, L15, L10, l5, L1}
        ]
        \addplot [
                  matrix plot,
                  point meta=explicit,
                  draw=gray!60,
                  line width=0.2pt
                ]
          coordinates {
          
            (0, 0) [0.9932727813720703] (1, 0) [0.9186698794364929] (2, 0) [0.9015313982963562] (3, 0) [0.9960143566131592] (4, 0) [0.9746895432472229] (5, 0) [0.9604776501655579] (6, 0) [0.9509255886077881] (7, 0) [0.9752033948898315] (8, 0) [0.99210524559021] 
    
            (0, 1) [0.8435842990875244] (1, 1) [0.9540755152702332] (2, 1) [0.9239829182624817] (3, 1) [0.9572973251342773] (4, 1) [0.9240286946296692] (5, 1) [0.9562387466430664] (6, 1) [0.9345027804374695] (7, 1) [0.9182559847831726] (8, 1) [0.9457600712776184] 
            
            (0, 2) [0.9025444388389587] (1, 2) [0.9604638814926147] (2, 2) [0.9711026549339294] (3, 2) [0.9732009768486023] (4, 2) [0.9310935139656067] (5, 2) [0.9396611452102661] (6, 2) [0.9750344157218933] (7, 2) [0.9445210695266724] (8, 2) [0.9795234203338623] 
            
            (0, 3) [0.9517195224761963] (1, 3) [0.9615983366966248] (2, 3) [0.9721279740333557] (3, 3) [0.9851235747337341] (4, 3) [0.9585769176483154] (5, 3) [0.9571506381034851] (6, 3) [0.9810774922370911] (7, 3) [0.9746212363243103] (8, 3) [0.9833089709281921] 
            
            (0, 4) [0.9663646817207336] (1, 4) [0.972140371799469] (2, 4) [0.9554597735404968] (3, 4) [0.9809578657150269] (4, 4) [0.9735614061355591] (5, 4) [0.9521092176437378] (6, 4) [0.9736505746841431] (7, 4) [0.9702660441398621] (8, 4) [0.973383903503418] 
            
            (0, 5) [0.9752957224845886] (1, 5) [0.9486376047134399] (2, 5) [0.9757123589515686] (3, 5) [0.9721248149871826] (4, 5) [0.9290626645088196] (5, 5) [0.9352152943611145] (6, 5) [0.9664879441261292] (7, 5) [0.9629886746406555] (8, 5) [0.9689842462539673] 
            
            (0, 6) [0.9496504664421082] (1, 6) [0.9669564366340637] (2, 6) [0.9823920130729675] (3, 6) [0.9774975180625916] (4, 6) [0.9552583694458008] (5, 6) [0.9730522036552429] (6, 6) [0.9710143804550171] (7, 6) [0.9712378978729248] (8, 6) [0.9849608540534973] 
            
            (0, 7) [0.9686731696128845] (1, 7) [0.9721410274505615] (2, 7) [0.9643929600715637] (3, 7) [0.9744583368301392] (4, 7) [0.9410427212715149] (5, 7) [0.9545309543609619] (6, 7) [0.979656994342804] (7, 7) [0.9543581604957581] (8, 7) [0.9750246405601501] 
            
            (0, 8) [0.9417765140533447] (1, 8) [0.9666762948036194] (2, 8) [0.9382836818695068] (3, 8) [0.9447276592254639] (4, 8) [0.9648952484130859] (5, 8) [0.9519550800323486] (6, 8) [0.9315083622932434] (7, 8) [0.9636140465736389] (8, 8) [0.9368050694465637] 
            
            (0, 9) [0.8887953758239746] (1, 9) [0.9232205152511597] (2, 9) [0.8180757761001587] (3, 9) [0.9040460586547852] (4, 9) [0.8172246217727661] (5, 9) [0.9062038660049438] (6, 9) [0.8409030437469482] (7, 9) [0.8453638553619385] (8, 9) [0.937285840511322] 
            
            (0, 10) [0.9567397236824036] (1, 10) [0.9373559951782227] (2, 10) [0.9842174053192139] (3, 10) [0.9651868939399719] (4, 10) [0.9647725820541382] (5, 10) [0.9628389477729797] (6, 10) [0.9813715815544128] (7, 10) [0.8873549103736877] (8, 10) [0.9828923940658569] 
            
            (0, 11) [0.825873851776123] (1, 11) [0.8540574908256531] (2, 11) [0.8655716776847839] (3, 11) [0.8953759670257568] (4, 11) [0.6034231185913086] (5, 11) [0.8764433860778809] (6, 11) [0.8577477931976318] (7, 11) [0.8418247103691101] (8, 11) [0.8924941420555115] 
            
            (0, 12) [0.6716381907463074] (1, 12) [0.5429750680923462] (2, 12) [0.5189183354377747] (3, 12) [0.7017351984977722] (4, 12) [0.5361546874046326] (5, 12) [0.8182823061943054] (6, 12) [0.6221371293067932] (7, 12) [0.7729945778846741] (8, 12) [0.5337312817573547] 
            
            (0, 13) [0.9558777213096619] (1, 13) [0.9653032422065735] (2, 13) [0.9459282159805298] (3, 13) [0.9507958889007568] (4, 13) [0.958426833152771] (5, 13) [0.9466043710708618] (6, 13) [0.9622576236724854] (7, 13) [0.9598395824432373] (8, 13) [0.9483599662780762] 
            
            (0, 14) [0.8747673630714417] (1, 14) [0.8632165193557739] (2, 14) [0.9209545254707336] (3, 14) [0.8936733603477478] (4, 14) [0.8878735303878784] (5, 14) [0.8521231412887573] (6, 14) [0.9055363535881042] (7, 14) [0.9151428937911987] (8, 14) [0.8848024606704712] 
            
            (0, 15) [0.9828934073448181] (1, 15) [0.976493239402771] (2, 15) [0.9908103942871094] (3, 15) [0.9930863380432129] (4, 15) [0.9883497953414917] (5, 15) [0.9859463572502136] (6, 15) [0.9898159503936768] (7, 15) [0.9778084754943848] (8, 15) [0.9881404638290405] 
            
            (0, 16) [0.9288504123687744] (1, 16) [0.9757133722305298] (2, 16) [0.9891662001609802] (3, 16) [0.9325534105300903] (4, 16) [0.9642255306243896] (5, 16) [0.9861147403717041] (6, 16) [0.9602741003036499] (7, 16) [0.9717326164245605] (8, 16) [0.9641994833946228] 
            
            (0, 17) [0.9783235788345337] (1, 17) [0.9676430821418762] (2, 17) [0.9825237393379211] (3, 17) [0.9906986355781555] (4, 17) [0.9927610158920288] (5, 17) [0.9830088019371033] (6, 17) [0.9865555763244629] (7, 17) [0.9729446172714233] (8, 17) [0.986799418926239] 
            
            (0, 18) [0.9996838569641113] (1, 18) [0.9998974800109863] (2, 18) [0.9998471736907959] (3, 18) [0.9998892545700073] (4, 18) [0.9999040365219116] (5, 18) [0.9998993873596191] (6, 18) [0.9998863935470581] (7, 18) [0.9998877048492432] (8, 18) [0.9996738433837891] 
            
            (0, 19) [0.9987314343452454] (1, 19) [0.9994142055511475] (2, 19) [0.999489426612854] (3, 19) [0.9995453953742981] (4, 19) [0.9994692206382751] (5, 19) [0.9992365837097168] (6, 19) [0.9992876648902893] (7, 19) [0.9991100430488586] (8, 19) [0.9995614886283875] 
            
            (0, 20) [0.9994315505027771] (1, 20) [0.9993491768836975] (2, 20) [0.9997063279151917] (3, 20) [0.9995618462562561] (4, 20) [0.9999024868011475] (5, 20) [0.9992208480834961] (6, 20) [0.9994626641273499] (7, 20) [0.9996819496154785] (8, 20) [0.9996793270111084] 
            
            (0, 21) [0.9548721313476562] (1, 21) [0.9658229351043701] (2, 21) [0.9962792992591858] (3, 21) [0.9637991189956665] (4, 21) [0.9787823557853699] (5, 21) [0.9736141562461853] (6, 21) [0.9494671821594238] (7, 21) [0.9505348801612854] (8, 21) [0.9973058700561523] 
            
            (0, 22) [0.9793946146965027] (1, 22) [0.9888956546783447] (2, 22) [0.9994254112243652] (3, 22) [0.9830456972122192] (4, 22) [0.984721839427948] (5, 22) [0.9884036183357239] (6, 22) [0.9813337922096252] (7, 22) [0.9948039650917053] (8, 22) [0.9886468052864075] 
            
            (0, 23) [0.9980415105819702] (1, 23) [0.9983205199241638] (2, 23) [0.9972345232963562] (3, 23) [0.9973574280738831] (4, 23) [0.9986960291862488] (5, 23) [0.9969883561134338] (6, 23) [0.9962148070335388] (7, 23) [0.9954519867897034] (8, 23) [0.9977066516876221] 
            
            (0, 24) [0.9975079298019409] (1, 24) [0.9976971745491028] (2, 24) [0.9975625276565552] (3, 24) [0.9967166781425476] (4, 24) [0.9979051351547241] (5, 24) [0.9977837204933167] (6, 24) [0.9977887868881226] (7, 24) [0.9978606104850769] (8, 24) [0.9973652958869934] 
            
            (0, 25) [0.9894563555717468] (1, 25) [0.9950647950172424] (2, 25) [0.9953436255455017] (3, 25) [0.9917457699775696] (4, 25) [0.9950376152992249] (5, 25) [0.9881582856178284] (6, 25) [0.9950093030929565] (7, 25) [0.9939633011817932] (8, 25) [0.9920754432678223] 
          };
    
      \end{axis}
      \draw [->, thick, red] 
            (-.8cm, 2.63cm) -- (-0.3cm, 2.63cm)  
            node [left, align=right, font=\small, rotate=90] at (-1cm, 3.55cm) {Intervention};
    \end{tikzpicture}
    }
  \end{center}
  \caption{Maximum probability values extracted from the attention head show a clear drop in confidence (used here as a proxy for semantic consistency and fluency) immediately after the layer where the intervention occurs.}
  \label{fig:heatmap}

\end{figure}

%% file: sections/signals.tex
\section{Mechanistic Signals}
\label{signals}
\subsection{Normalized Branching Factor}

\label{nbf}
Let $P_{\theta}^{(\mathrm{eff})}(y_t \mid X, y_{<t}) \;=\; \mathrm{softmax}\!\big(z_t\!\restriction_{V_{\mathrm{eff}}(t)}\big)$ denote the effective vocabulary probablity distribution at generation step $t$. 
We define the \emph{branching factor} at time step $t$ as the exponential of the entropy of the output distribution,
\[
B_t \;=\; \exp\!\big( H(p_t^{(\mathrm{eff})}) \big)
\]
where the entropy of any given distribution is calcualted as 
\[
H(p) = - \sum_{y = 1}^{N} p(y)\log p(y)
\]
To account for sequence length effects, we define the \emph{normalized branching factor} up to time step $T$ as \[ \bar{B}_{1:T} \;=\; \frac{1}{T} \sum_{t=1}^{T} B_t. \]

Figures \ref{fig:good_nbf} and \ref{fig:bad_nbf} present the \ac{nbf} signals across the generation steps in the final layer of the \ac{llm}.

\subsection{\ac{kl} Divergence difference}

Let $h_t^{(\ell)}$ and $\hat{h}_t^{(\ell)}$ denote the residual representations before and after the steering intervention at layer $\ell$ and time step $t$. Using the model’s unembedding function as defined in Equation~\ref{eq:unembed}, residual representations are mapped to effective vocabulary distributions in a manner analogous to logit-lens–based interpretations \cite{wang2025logitlens4llms}:
\[
p_t^{(\ell)} = \mathrm{U}\!\left(h_t^{(\ell)}\right),
\qquad
\hat{p}_t^{(\ell)} = \mathrm{U}\!\left(\hat{h}_t^{(\ell)}\right).
\]

Similarly, the effective vocabulary distribution induced by the steering vector $s$ is defined as
\[
q^{(\ell)} = \mathrm{U}\!\left(s\right),
\]
which remains invariant across time steps in this setting.

The \emph{\ac{kl} difference} induced by steering at layer $\ell$ and time step $t$ is then defined as
\[
\mathrm{Diff}_t^{(\ell)} 
\;=\;
\mathrm{KL}\!\left(p_t^{(\ell)} \,\|\, q^{(\ell)}\right)
\;-\;
\mathrm{KL}\!\left(\hat{p}_t^{(\ell)} \,\|\, q^{(\ell)}\right).
\]

Figures \ref{fig:good_kl} and \ref{fig:bad_kl} depict the \ac{kl} signals across the generation steps in the 12th layer of the \ac{llm}.

\input{graphs/KL_9b_Christian_addition_act_13}

\subsection{Attention Pattern Structure}

Let $Attn_t^{(\ell)} \in \mathbb{R}^{t \times t}$ (as it is defined in \ref{eq:attention}) denote the self-attention probability matrix at layer $\ell$ and time step $t$. We consider the layer $\ell^{+}$ immediately following the steering intervention as the primary site for extracting attention pattern signals. Figure \ref{fig:heatmap} shows a clear pattern of attention disruption after intervention.

%% file: graphs/KL_9b_Christian_addition_act_13.tex
\begin{figure}[ht]
  \vskip 0.2in
  \begin{center}
    \resizebox{\columnwidth}{!}{%

\begin{tikzpicture}
\begin{axis}[
    width=14cm,
    height=5.05cm,
    grid=both,
    xmin=1,
    xmax=30,
    ymin=11.25,
    ymax=12.75,
    xlabel={Generation step},
    ylabel={Measured quantity},
    legend style={
        at={(-0.05,0.5)},
        anchor=east
    },
    yticklabel pos=right,
    ylabel=\ac{kl} divergence, 
    ylabel near ticks,
    legend columns=1,
    legend cell align=left,
    cycle list name=exotic,
]

\addplot [color=blue!60!black, thick] coordinates {
(1, 11.5) (2, 12.18) (3, 12.01) (4, 12.01) (5, 12.11) (6, 12.08) (7, 12.02) (8, 12.01) (9, 11.93) (10, 11.98) (11, 11.91) (12, 11.85) (13, 11.89) (14, 11.88) (15, 11.87) (16, 11.97) (17, 11.89) (18, 11.86) (19, 11.84) (20, 11.86) (21, 11.88) (22, 11.95) (23, 11.94) (24, 11.92) (25, 11.91) (26, 11.91) (27, 11.92) (28, 11.89) (29, 11.93) (30, 11.84)
};
\addlegendentry{$Diff_{unsteered}$}

\addplot [color=red!60!black, thick] coordinates {
(1, 11.26) (2, 11.77) (3, 11.83) (4, 11.87) (5, 11.89) (6, 11.89) (7, 11.89) (8, 11.89) (9, 11.93) (10, 11.91) (11, 11.96) (12, 11.93) (13, 11.92) (14, 11.97) (15, 11.93) (16, 11.94) (17, 11.94) (18, 11.94) (19, 11.91) (20, 11.94) (21, 11.92) (22, 11.92) (23, 11.93) (24, 11.94) (25, 11.91) (26, 11.9) (27, 11.91) (28, 11.88) (29, 11.92) (30, 11.91)
};
\addlegendentry{$Diff_{steered}$}

\addplot+[domain=1:30, dashed, blue!60!black] {11.923377018235623};
\addlegendentry{Mean $Diff_{unsteered}$}

\addplot+[domain=1:30, dashed, red!60!black] {11.888146032889685};
\addlegendentry{Mean $Diff_{steered}$}

\end{axis}
\end{tikzpicture}
    }
  \end{center}
  \caption{Unsuccessful steering, as evidenced by the lack of a significant difference in \ac{kl} divergence between the steered and unsteered representations.}
  \label{fig:bad_kl}
\end{figure}
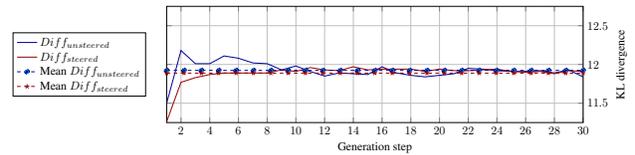

%% file: graphs/KL_9b_Christian_rotation_act_5.tex
\begin{figure}[ht]
  \vskip 0.2in
  \begin{center}
    \resizebox{\columnwidth}{!}{%

\begin{tikzpicture}
\begin{axis}[
    width=14cm,
    height=5.05cm,
    grid=both,
    xmin=1,
    xmax=30,
    ymin=11.25,
    ymax=12.75,
    xlabel={Generation step},
    ylabel={Measured quantity},
    legend style={
        at={(-0.05,0.5)},
        anchor=east
    },
    yticklabel pos=right,
    ylabel=\ac{kl} divergence, 
    ylabel near ticks,
    legend columns=1,
    legend cell align=left,
    cycle list name=exotic,
]

\addplot [color=blue!60!black, thick] coordinates {
(1, 11.5) (2, 12.23) (3, 12.15) (4, 12.12) (5, 12.1) (6, 12.03) (7, 12.03) (8, 11.97) (9, 11.89) (10, 11.96) (11, 11.91) (12, 11.9) (13, 11.86) (14, 11.83) (15, 11.93) (16, 11.87) (17, 11.83) (18, 11.84) (19, 11.83) (20, 11.83) (21, 11.86) (22, 11.75) (23, 11.68) (24, 11.84) (25, 11.79) (26, 11.73) (27, 11.82) (28, 11.73) (29, 11.8) (30, 11.77)
};
\addlegendentry{$Diff_{unsteered}$}

\addplot [color=red!60!black, thick] coordinates {
(1, 11.54) (2, 12.0) (3, 11.97) (4, 11.86) (5, 11.83) (6, 11.71) (7, 11.58) (8, 11.63) (9, 11.56) (10, 11.49) (11, 11.49) (12, 11.46) (13, 11.5) (14, 11.46) (15, 11.52) (16, 11.55) (17, 11.37) (18, 11.44) (19, 11.3) (20, 11.36) (21, 11.33) (22, 11.3) (23, 11.36) (24, 11.33) (25, 11.26) (26, 11.4) (27, 11.28) (28, 11.35) (29, 11.32) (30, 11.33)
};
\addlegendentry{$Diff_{steered}$}

\addplot+[domain=1:30, dashed, blue!60!black] {11.878892929231126};
\addlegendentry{Mean $Diff_{unsteered}$}

\addplot+[domain=1:30, dashed, red!60!black] {11.49521914968888};
\addlegendentry{Mean $Diff_{steered}$}

\end{axis}
\end{tikzpicture}
    }
  \end{center}
  \caption{Successful steering, as evidenced by a significant difference in KL divergence between the steered and unsteered representations.}
  \label{fig:good_kl}
\end{figure}
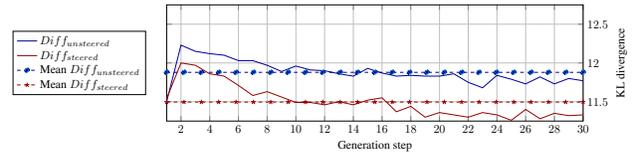

%% file: sections/method.tex
\section{Methodology}
\label{method}

Based on the notation introduced in Section~\ref{formalization} and following the experimental setups proposed by \cite{sooInterpretableSteeringLarge2025,turnerSteeringLanguageModels2024a}, this research extracts three sets of mechanistic signals for probing steering quality. Owing to the qualitative nature of steering-quality assessment—and to manage annotation costs—we treat \ac{llm}-generated annotations as proxies for human expert annotations. This choice is validated through an inter-judge agreement analysis conducted prior to the main experiment.

The main experiment involves training a regression model on the pre-extracted signals and evaluating its performance on a held-out set of unseen experiments. We ensure that no experiment appears in both the training and testing sets. Figure \ref{fig:pipeline} depicts the feature extraction and regression phases as inputs pass through the model.

%% file: sections/experiments.tex
\section{Experimental Design}
\label{experiments}
\subsection{Models and Tasks}

Experiments are conducted on the \textsc{Gemma-2} model family for methodological rather than performance-driven reasons, as the analysis requires full access to intermediate residual representations and compatibility with publicly available SAE feature dictionaries.

All evaluations consider a \emph{free-form text generation} task initiated from a neutral prompt, \emph{``I think ...''}, selected to minimize prior bias toward any specific attribute. A set of $|C| = 9$ distinct steering concepts is studied. For each concept $c \in C$, a corresponding steering vector $s$ is extracted using two widely adopted methods: \ac{caa} and \ac{sae}. These vectors serve as inputs to the residual steering functions defined in Equation~\ref{eq:steering_fnc}.

Two variants of steering functions are evaluated: additive steering (Section~\ref{fnc:add}) and rotation-based steering (Section~\ref{fnc:rot}). Following prior work \cite{sooInterpretableSteeringLarge2025, turnerSteeringLanguageModels2024a}, steering is applied at layer $\mathcal{L} = \{12\}$, with the steering strength varied over $\alpha \in \{0, 20, 40, \dots, 320\}$.

To assess steering effectiveness at the behavioral level, \gpt \space and \gmn \space are employed as independent evaluators to score generated outputs with respect to concept alignment and overall coherence, using a fixed evaluation prompt adopted from \cite{sooInterpretableSteeringLarge2025}. These external evaluations are used solely for behavioral validation and complement the internal mechanistic metrics. In total, 2{,}304 experimental runs are conducted, covering all combinations of models, concepts, steering methods, and steering strength parameters.




\input{graphs/steeringfunction}

\subsection{Residual Steering Functions}
\label{steering_functions}
Let $h_t^{(\ell)} \in \mathbb{R}^d$ denote the residual representation at layer $\ell$ and time step $t$, and let $s \in \mathbb{R}^d$ be a steering vector. We define two steering functions $F(\cdot)$ operating directly on the residual stream.

\subsubsection{Residual Addition Steering}

\label{fnc:add}
The simplest form of steering is additive intervention in the residual space. We define the additive steering function for a scaling coefficient $\alpha \in \mathbb{R}$ as
\[
\tilde{h}_t^{(\ell)} 
\;=\;
F_{\mathrm{add}}\!\left(h_t^{(\ell)}, s, \alpha\right)
\;=\;
h_t^{(\ell)} + \alpha s.
\]

This operation applies a linear shift of the residual representation in the direction of the steering vector, without preserving the original residual norm.

\label{fnc:add}


\subsubsection{Residual Rotation Steering}
\label{fnc:rot}
To preserve the magnitude of the residual stream while modifying its direction, we define a rotation-based steering function using a tangent-space exponential map.

Let
\[
\hat{x} = \frac{h_t^{(\ell)}}{\|h_t^{(\ell)}\|}, 
\qquad
\hat{y} = \frac{s}{\|s\|}
\]
denote the normalized residual and steering directions, respectively. The angle between them is
\[
\theta = \arccos\!\left(\hat{x}^{\top} \hat{y}\right).
\]

We compute the component of $\hat{y}$ orthogonal to $\hat{x}$ as
\[
v = \hat{y} - (\hat{x}^{\top}\hat{y})\hat{x},
\qquad
\hat{v} = \frac{v}{\|v\|}.
\]

Given a steering strength $\beta = \alpha/320, \beta \in [0,1]$, we define the rotation angle
\[
\phi = \beta \theta.
\]

The steered residual direction is obtained via a rotation in the two-dimensional subspace spanned by $\hat{x}$ and $\hat{v}$:
\[
\hat{z} = \cos(\phi)\,\hat{x} + \sin(\phi)\,\hat{v}.
\]

Finally, the original residual norm is restored:
\[
\tilde{h}_t^{(\ell)} 
\;=\;
F_{\mathrm{rot}}\!\left(h_t^{(\ell)}, s, \alpha\right)
\;=\;
\|h_t^{(\ell)}\|\,\hat{z}.
\]

This intervention preserves the magnitude of the residual stream while smoothly rotating its direction toward the steering vector. It can be interpreted as a geodesic update on the unit hypersphere, preventing norm inflation and reducing unintended entropy collapse.

%% file: graphs/steeringfunction.tex
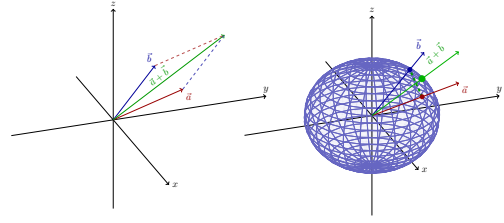
\begin{figure}[ht]
  \vskip 0.2in
  \begin{center}
    \resizebox{0.8\columnwidth}{!}{%

\begin{tikzpicture}
\begin{scope}[xshift=-8cm]
\begin{axis}[
    view={70}{30},
    axis lines=center,
    axis line style={->, thick},
    xlabel=$x$,
    ylabel=$y$,
    zlabel=$z$,
    xlabel style={anchor=west},
    ylabel style={anchor=south},
    zlabel style={anchor=south},
    xmin=-2, xmax=3,
    ymin=-2, ymax=3,
    zmin=-2, zmax=2.5,  
    grid=major,
    width=14cm,
    height=14cm,
    ticks=none,
]

\addplot3[
    quiver={
        u=1, v=1, w=1,
        scale arrows=1,
        every arrow/.append style={->, red!60!black, thick}
    },
    -stealth,
] coordinates {(0,0,0)};
\node[red!60!black, below right] at (axis cs:1,1,1) {$\vec{a}$};

\addplot3[
    quiver={
        u=-0.5, v=1, w=0.8,
        scale arrows=1,
        every arrow/.append style={->, blue!60!black, thick}
    },
    -stealth,
] coordinates {(0,0,0)};
\node[blue!60!black, above left] at (axis cs:-0.5,1,0.8) {$\vec{b}$};

\addplot3[
    quiver={
        u=0.5, v=2, w=1.8,
        scale arrows=1,
        every arrow/.append style={->, green!60!black, thick}
    },
    -stealth,
] coordinates {(0,0,0)};
\node[green!60!black, inner sep=5pt, rotate=38] at (axis cs:0.5, 0.7, 1.1) {$\vec{a} + \vec{b}$};

\addplot3[
    dashed,
    thick,
    blue!60!black!70
] coordinates {(1,1,1) (0.5,2,1.8)};

\addplot3[
    dashed,
    thick,
    red!60!black!70
] coordinates {(-0.5,1,0.8) (0.5,2,1.8)};

\end{axis}
\end{scope}

\begin{axis}[
    view={70}{30},
    axis lines=center,
    axis line style={->, thick},
    xlabel=$x$,
    ylabel=$y$,
    zlabel=$z$,
    xlabel style={anchor=west},
    ylabel style={anchor=south},
    zlabel style={anchor=south},
    xmin=-2, xmax=2,
    ymin=-2, ymax=2,
    zmin=-2, zmax=2,
    grid=major,
    width=14cm,
    height=14cm,
    ticks=none,
]

\addplot3[
    surf,
    opacity=0.02,
    colormap={blue!60!black}{color(0cm)=(blue!60!black); color(1cm)=(blue!60!black)},
    samples=30,
    domain=0:360,
    y domain=0:180,
    z buffer=sort,
    faceted color=none,
] ({cos(x)*sin(y)}, {sin(x)*sin(y)}, {cos(y)});

\addplot3[
    mesh,
    draw=blue!60!black!60,
    line width=0.05pt,
    samples=25,
    domain=0:360,
    y domain=0:180,
] ({cos(x)*sin(y)}, {sin(x)*sin(y)}, {cos(y)});

\addplot3[
    quiver={
        u=1, v=1, w=1,
        scale arrows=1,
        every arrow/.append style={->, red!60!black, thick}
    },
    -stealth,
] coordinates {(0,0,0)};
\node[red!60!black, below right] at (axis cs:1,1,1) {$\vec{a}$};

\addplot3[
    quiver={
        u=-0.5, v=1, w=0.8,
        scale arrows=1,
        every arrow/.append style={->, blue!60!black, thick}
    },
    -stealth,
] coordinates {(0,0,0)};
\node[blue!60!black, above left] at (axis cs:-0.5,1,0.8) {$\vec{b}$};

\addplot3[
    green!60!black,
    dashed,
    line width=2pt,
]
coordinates {
    ({1/sqrt(3)}, {1/sqrt(3)}, {1/sqrt(3)})
    ({-0.5/sqrt(1.89)}, {1/sqrt(1.89)}, {0.8/sqrt(1.89)})
};

\addplot3[
    only marks,
    mark=*,
    mark size=2pt,
    red!60!black
] coordinates {
    ({1/sqrt(3)}, {1/sqrt(3)}, {1/sqrt(3)})
};

\addplot3[
    only marks,
    mark=*,
    mark size=2pt,
    blue!60!black
] coordinates {
    ({-0.5/sqrt(1.89)}, {1/sqrt(1.89)}, {0.8/sqrt(1.89)})
};

\pgfmathsetmacro{\ax}{1/sqrt(3)}
\pgfmathsetmacro{\ay}{1/sqrt(3)}
\pgfmathsetmacro{\az}{1/sqrt(3)}

\pgfmathsetmacro{\bnorm}{sqrt(1.89)}
\pgfmathsetmacro{\bx}{-0.5/\bnorm}
\pgfmathsetmacro{\by}{1/\bnorm}
\pgfmathsetmacro{\bz}{0.8/\bnorm}

\pgfmathsetmacro{\sx}{\ax+\bx}
\pgfmathsetmacro{\sy}{\ay+\by}
\pgfmathsetmacro{\sz}{\az+\bz}

\pgfmathsetmacro{\snorm}{sqrt(\sx*\sx+\sy*\sy+\sz*\sz)}
\pgfmathsetmacro{\cx}{\sx/\snorm}
\pgfmathsetmacro{\cy}{\sy/\snorm}
\pgfmathsetmacro{\cz}{\sz/\snorm}

\pgfmathsetmacro{\scale}{sqrt(3)}
\addplot3[
    quiver={
        u={\scale*\cx},
        v={\scale*\cy},
        w={\scale*\cz},
        scale arrows=1,
        every arrow/.append style={->, green!70!black, thick}
    },
    -stealth,
] coordinates {(0,0,0)}; 

\addplot3[
    only marks,
    mark=*,
    mark size=3pt,
    green!70!black
]
coordinates {(\cx,\cy,\cz)}
node[green!60!black, inner sep=5pt, rotate=45] at (axis cs:0.5, 0.8, 1.35) {$\vec{a} + \vec{b}$};
\end{axis}
\end{tikzpicture}
    }
  \end{center}
  \caption{Comparison of the naïve additive steering function (left) with the proposed rotational steering function (right), which theoretically accounts for preserving the magnitude of the residual representation.}

\end{figure}

%% file: sections/results.tex
\section{Results}
\label{results}
\input{tabels/regression}
\subsection{Reliability of LLMs for Qualitative Assessment}
\label{reliability}

Since the regression experiments in this setting rely exclusively on annotations provided by \ac{llm} judges, the reliability of \ac{llm}-based qualitative steering assessment is evaluated first. Accordingly, \gpt \space is selected as the primary evaluator, with \gmn \space included as an independent and comparably capable complementary judge.

The evaluation dataset comprises 72 steering conditions, spanning 9 steering concepts, 2 steering vector extraction methods, 2 steering functions, and 2 base language models. Each generated output is independently scored by both evaluators using an identical evaluation prompt.

Inter-rater reliability is quantified using the intraclass correlation coefficient under a fixed-raters, absolute-agreement formulation (ICC(3,1)). The resulting agreement score is
\[
\mathrm{ICC}(3,1) = 0.78,
\quad 95\%~\mathrm{CI} = [0.67, 0.86],
\]
with $F(71,71) = 8.02$ and $p < 10^{-15}$, indicating substantial agreement between the two \ac{llm} judges.

Agreement metrics sensitive to scale and offset, such as Krippendorff’s $\alpha$, yield a lower value of $\alpha = 0.23$ when computed on raw scores, despite a high Pearson correlation of $r = 0.84$. This discrepancy reflects \emph{systematic calibration differences} rather than semantic disagreement. In particular, \gmn \space consistently assigns higher absolute scores, with a mean of $0.40$, compared to $0.22$ for \gpt.

To correct for this calibration bias, per-judge z-score normalization is applied prior to recomputing agreement metrics. After normalization, Krippendorff’s $\alpha$ increases substantially to
\[
\alpha_{\text{z-scored}} = 0.85,
\]
indicating strong agreement in relative qualitative judgments between the two evaluators.


\subsection{Entropy and KL Dynamics Under Steering}
\label{entropy_kl}
\ac{nbf} is used as a proxy for the model’s effective generative capacity during decoding, or as an indicator of entropy collapse, which should be preserved—and in some cases increase—with higher steering intensity. Figure~\ref{fig:good_nbf} illustrates a case of effective steering, showing the change in \ac{nbf} across different steering strengths when the \emph{\ac{sae}} extraction method is applied to the \emph{Gemma~2–2B} model for the concept \emph{London} using the \emph{rotation} steering function. In contrast, Figure~\ref{fig:bad_nbf} presents an ineffective steering case under an identical setting, differing only in the steering function, which is \emph{addition}.

Such increases in entropy may arise either from meaningful redistribution of probability mass aligned with the steering signal or from degenerate flattening of the output distribution. Distinguishing between these two regimes is therefore essential. To disambiguate these effects, the dynamics of \ac{kl} divergence between steered and unsteered residual representations are examined. \ac{kl} can serve as an indicator for proximity to the vocabulary distribution induced directly by the steering vector is measured.

A reduction in \ac{kl} divergence toward the steering-induced distribution indicates structured, concept-aligned redistribution (Figure~\ref{fig:good_kl}). This behavior is observed in an experiment using \emph{Gemma~2–9B} steered toward the concept \emph{Christianity}, with \ac{caa} extraction and the \emph{rotation} steering function at $\alpha = 100$ (score = 0.22). In contrast, entropy increases without corresponding \ac{kl} shifts are indicative of non-informative flattening (Figure~\ref{fig:bad_kl}). This regime appears under otherwise identical conditions—model, concept, and extraction method—except that the steering function is \emph{addition} and the steering strength is increased to $\alpha = 260$ (score = 0.11). These examples are intentionally selected, as they exhibit nearly identical \ac{nbf} trajectories (appendix section \ref{similar_different}) despite fundamentally different underlying dynamics.

Another notable result is the strong correlation between the language fluency metric $\mathcal{C}(\mathcal{T})$ and the maximum attention-head probability extracted immediately after the intervention position. The observed correlations are 0.71 and 0.72 for the \gpt \space and \gmn \space judges, respectively. In contrast, no comparable correlation is observed in later layers. Figure~\ref{fig:heatmap} illustrates this pattern for the \emph{Gemma~2–2B} model across all nine steering concepts.

\subsection{Predicting Steering Quality from Internal Signals}

To evaluate whether the information required to predict steered generation quality is present in the aforementioned mechanistic signals (\ref{signals}), a regression-based analysis is conducted that maps internal diagnostics to external quality assessments (Figure~\ref{fig:pipeline}). This experiment directly tests the main hypothesis.

For each steering configuration, a feature vector is extracted comprising the steering strength $\alpha$, the \ac{nbf}, and \ac{kl} divergence metrics computed before and after steering. All combinations of 16 steering strengths, 2 steering vector extraction methods, 2 steering functions, 2 base models, and 9 steering concepts are considered, resulting in a total of 1{,}152 experimental conditions.

Using these features, a regression model is trained to predict the combined performance metric\footnote{Referred to throughout this paper as the steering performance or steering score.}. For further details, please refer to Appendix~\ref{ex_details}.

\[
P(\mathcal{T}) = \mathcal{S}(\mathcal{T}) \times \mathcal{C}(\mathcal{T}).
\]
Evaluation is performed using three independent random seeds with a 70/30 train–test split to ensure robustness across steering configurations.

To evaluate the dependence on the choice of judge, the experiment was repeated using qualitative scores obtained independently from both \gpt \space and \gmn \space as supervision signals. The predictive performance of the resulting regression models for each judge is reported in Table~\ref{tab:regression}. The distribution of predicted values and ground truth annotations for seed number 10 is provided in Section~\ref{regression_overview} for further details.






\input{tabels/baseline}

\subsection{Comparison with Stronger Baselines}

Analysis of the internal dynamics associated with successful steering—as characterized by \ac{nbf} preservation, meaningful \ac{kl} divergence shifts, and stable attention patterns—reveals several limitations of naive residual addition.

First, additive steering implicitly assumes linearity in the residual space: it treats the effect of a steering vector as independent of the current residual direction. This assumption is poorly aligned with the highly nonlinear geometry induced by layer normalization and attention mechanisms. As a result, linear addition can induce disproportionate changes in token distributions, often leading to entropy collapse or attention instability.

Second, residual addition disregards the magnitude of the original residual representation. Because the intervention applies a fixed shift $\alpha s$ regardless of $\|h_t^{(\ell)}\|$, the relative influence of the steering vector varies unpredictably across layers, time steps, and tokens. This sensitivity can amplify small steering signals or overwhelm large residuals, degrading coherence.

To address these limitations, we propose rotation-based steering in \ref{fnc:rot}, which modifies only the direction of the residual representation while explicitly preserving its norm. By operating on the unit hypersphere via a geodesic update, rotation steering respects the local geometry of the residual stream and produces controlled interventions.

Empirically, we find that rotation steering achieves stronger alignment with the desired control signal while better preserving entropy, \ac{kl} dynamics, and attention stability (table \ref{tab:baseline}). Notably, rotation steering can reuse the same steering vectors as additive methods, but applies them in a more effective and robust manner, resulting in improved steering–coherence trade-offs. 

%% file: tabels/regression.tex
\begin{table}[t]
  \caption{Regression model performance across different \ac{llm} annotators and evaluation metrics.}
  \label{tab:regression}
  \begin{center}
    \begin{small}
      \begin{sc}
        \begin{tabular}{lccc}
          \toprule
          Metric  & \gpt         & \gmn     \\
          \midrule
          MAE    & 0.0531 ± 0.0000 & 0.0871 ± 0.0000 \\
          RMSE & 0.0728 ± 0.0000 & 0.1160 ± 0.0001 \\
          R2    & 0.4698 ± 0.0022 & 0.5445 ± 0.0012 \\
          \bottomrule
        \end{tabular}
      \end{sc}
    \end{small}
  \end{center}
  \vskip -0.1in
\end{table}

%% file: tabels/baseline.tex
\begin{table}[t]
  \caption{Performance comparison of naïve additive steering as a common baseline versus rotational steering functions across different annotators.}
  \label{tab:baseline}
  \begin{center}
    \begin{small}
      \begin{sc}
        \begin{tabular}{llcccc}
\toprule
 &  & \multicolumn{2}{c}{ChatGPT} & \multicolumn{2}{c}{Gemini} \\

\textbf{Model} & \textbf{Method} & \textbf{Add} & \textbf{Rot} & \textbf{Add} & \textbf{Rot} \\
\midrule
\multirow{2}{*}{Gemma2-2b}
 & CAA & 0.22 & \textbf{0.28} & 0.40 & \textbf{0.46} \\

 & SAE & 0.13 & \textbf{0.16} & 0.26 & \textbf{0.32} \\
\midrule
\multirow{2}{*}{Gemma2-9b}
 & CAA & 0.25 & \textbf{0.29} & 0.45 & \textbf{0.52} \\

 & SAE & \textbf{0.22} & 0.21 & \textbf{0.40} & \textbf{0.40} \\
\bottomrule
        \end{tabular}
      \end{sc}
    \end{small}
  \end{center}
  \vskip -0.1in
\end{table}

%% file: sections/related.tex
\section{Related Work}
\label{related}

\noindent\textbf{Activation-based steering.} Activation-based steering alters \acp{llm} behavior at inference time by intervening in activation space rather than retraining. A widely used method is \ac{caa}, which derives steering directions from contrastive input pairs and has been shown to control behavior across many settings, including improved chess play \cite{karvonenEmergentWorldModels2024}, semantic concept steering \cite{sooInterpretableSteeringLarge2025, turner2023steering, wuAxBenchSteeringLLMs2025, sunHyperDASAutomatingMechanistic2025}, toxicity reduction \cite{nguyenMultiAttributeSteeringLanguage}, mitigation of hallucination and psychopathy-related tendencies \cite{rimsky-etal-2024-steering}, and stylistic personalization \cite{zhang-etal-2025-personalized}. \textbf{Interpretable steering with \acp{sae}.} A more interpretable line of work uses sparse autoencoders to decompose polysemantic residual activations into human-readable latent features, enabling steering along learned dimensions for tasks such as mathematical reasoning \cite{wangImprovingLLMReasoning2025}, concept-level control \cite{sooInterpretableSteeringLarge2025, aradSAEsAreGood2025, choCorrSteerSteeringImproves2025, chalnevImprovingSteeringVectors2024}, and multi-concept manipulation \cite{joshiIdentifiableSteeringSparse2025}. However, \ac{sae}-based approaches depend on a fixed learned feature dictionary, which may omit factors needed for reliable steering under diverse contexts. \textbf{Mechanistic understanding.} Recent studies analyze how contrastive steering vectors behave internally, characterizing their structure, generalization, and limitations \cite{haoPatternsMechanismsContrastive2025, tanAnalyzingGeneralizationReliability2025, chenSTEERBENCHBenchmarkEvaluating2025}, and highlighting failures caused by distribution shift and representational entanglement \cite{niranjanLimitationsSteeringLanguage2025}. These works provide valuable diagnostics but are mostly descriptive/post-hoc and do not connect steering success to principled mechanistic signals grounded in probabilistic generation or language-modeling theory. \textbf{Evaluation via LLM-as-judge.} Since human evaluation is expensive, many benchmarks use \acp{llm} as judges to approximate expert scoring \cite{wuAxBenchSteeringLLMs2025, sunHyperDASAutomatingMechanistic2025, sooInterpretableSteeringLarge2025, chalnevImprovingSteeringVectors2024}; yet judge calibration and reliability are often assumed rather than tested, leaving open issues around judge dependence, score stability, and robustness of qualitative conclusions.

%% file: sections/discussion.tex
\section{Discussion}
\label{discussion}
\subsection{Implications for Mechanistic Interpretability}


Modeling the steering operation within the general framework of reading from and writing to the residual stream suggests that mechanistic signals—including the entropy at each layer as a proxy for generative capacity, and the \ac{kl} divergence of the steered vocabulary as a proxy for distributional alignment—possess reasonable predictive power for assessing final steering quality without reliance on any external judge.

Based on results in Table \ref{tab:regression} The regression model exhibits consistent behavior across both label sets, indicating stable learning dynamics under identical experimental conditions. While performance differs in absolute error magnitude, the overall trends are coherent. Annotation produced by \gpt \space yields substantially lower MAE (0.0531 vs. 0.0871) and RMSE (0.0728 vs. 0.1160), suggesting tighter pointwise agreement and smaller residual dispersion. In contrast, annotation produced by \gmn \space achieves a higher coefficient of determination (R² = 0.5445 vs. 0.4698), indicating that although prediction errors are larger in scale, the model captures a greater proportion of variance in that annotation space. 

RMSE penalizes large errors more than MAE. The fact that RMSE is only moderately larger than MAE in both annotation sets suggests that there are no extreme outliers dominating the error, i.e., errors are relatively evenly distributed rather than having a few very large deviations. Overall, the model demonstrates robust generalization across annotators, with error profiles suggesting primarily scale-dependent discrepancies rather than structural prediction failures (similar to section \ref{reliability} discussion).

\subsection{Implications for Evaluation and Reliability}

As \acp{llm} are increasingly used for qualitative analysis and evaluation of generated text, there is a growing need for systematic methods to assess the reliability of their judgments. 

In Section~\ref{reliability}, we analyze the agreement between two affordable and comparably capable models, \gpt \space and \gmn \space, when applied to the same qualitative evaluation tasks. We observe a substantial level of agreement between these models, suggesting that, under controlled conditions, such models can serve as reasonable proxies for human evaluators.

%% file: sections/limits.tex
\section{Limitations}
\label{limits}

\paragraph{Limitations.}
While this study conducts a total of 2{,}304 experiments across diverse settings, it considers only nine target concepts. This limited conceptual coverage constrains the strength of the conclusions and may not fully reflect the variability of steering behavior across a broader semantic space. In addition, our analysis focuses exclusively on the \textsc{Gemma} model family, which prioritizes interpretability access and tool availability over alignment with the most recent frontier models. As a result, the findings may not directly generalize to newer and more capable models. Addressing this limitation is non-trivial, as the number of \ac{llm} families with \acp{sae} trained during the pretraining phase remains limited, yet conducting \ac{sae}-based analyses necessitates availability of such models.

From a theoretical perspective, the current regression formulation requires further refinement to improve its expressive capacity and better capture variance in the data. Moreover, label noise presents a significant challenge: it arises partly from reliance on \ac{llm}-based judges and partly from the inherently qualitative nature of the evaluated tasks. This noise complicates generalization and introduces additional uncertainty in the reported findings.



%% file: sections/conclusion.tex
\section{Conclusion}
\label{conclude}

In this work, steering is framed as a familiar read–write operation on the residual stream of an \ac{llm}, enabling its analysis using established toolkits from the mechanistic interpretability literature. Steering reliability is shown to be systematically evaluable through internal model analysis. Beyond post-hoc inspection, the results demonstrate that internal mechanistic signals can be used to predict steering effectiveness prior to full text generation. Moreover, under controlled experimental conditions, \acp{llm} are shown to serve as useful proxies for qualitative evaluators.

A mechanistic account of activation-based steering is provided through analysis of entropy and \ac{kl} divergence dynamics across layers and decoding steps. Effective steering is characterized by structured entropy preservation together with controlled \ac{kl} divergence behavior. These findings support an interpretation of steering as a controlled transformation of the model’s internal distribution, rather than as an arbitrary perturbation.

Finally, a stronger steering-function baseline is introduced enhancing two of the most widely used activation-based steering methods, \ac{caa} and \ac{sae}. This baseline is grounded in a principled theoretical formulation and demonstrates empirical improvements over naïve additive steering. By distinguishing meaningful steering directions from arbitrary perturbations. 

Taken together, these findings position steering as a measurable, predictable, and testable process, and highlight the value of internal model dynamics for both evaluation and interpretability. This work aims to motivate future research on mechanistically grounded approaches to controlling \ac{llm} behavior.

%% file: sections/impacts.tex
\section*{Broader Impact and Ethical Considerations}

This work investigates the mechanistic foundations of activation-based steering in large language models, with the aim of improving the predictability, reliability, and interpretability of inference-time control methods. By connecting behavioral steering outcomes to internal model dynamics, the proposed analysis framework may support safer deployment practices and reduce reliance on trial-and-error intervention strategies.

At the same time, steering techniques may be misused to amplify harmful, misleading, or manipulative behaviors if applied without appropriate safeguards. This work does not introduce new steering mechanisms or expand the expressive power of existing methods; rather, it provides diagnostic tools for analyzing and evaluating steering behaviors already present in current approaches. As such, the results should be interpreted as analytical insights rather than recommendations for deploying specific steering objectives.

All experiments are restricted to open-weight models and non-sensitive steering concepts. The proposed metrics characterize the stability and effectiveness of steering signals but do not assess the social desirability or ethical appropriateness of any particular steering direction. Determinations regarding acceptable model behavior remain application-dependent and outside the scope of this study.

The use of large language models as qualitative evaluators introduces potential concerns related to bias, calibration differences, and evaluator consistency. To address these issues, this work explicitly measures inter-judge reliability across architecturally distinct models and applies normalization procedures to account for systematic calibration offsets. The results indicate strong agreement in relative qualitative assessments, supporting the use of LLM-based judges as scalable—though imperfect—proxies for human evaluation.

Finally, this study does not involve any model training or fine-tuning. All analyses are conducted through inference-time interventions and forward-pass inspection, resulting in substantially lower computational and environmental costs compared to retraining-based alignment or adaptation methods.

%% file: sections/ack.tex
\section{Acknowledgement}
This research was supported by the ARC Centre of Excellence for Automated Decision-Making and Society (CE200100005). We also acknowledge ResetData and the National Computational Infrastructure (NCI) for providing the computational resources that enabled our experiments.

%% file: sections/app.tex
\section{Regression Setup and Feature Construction}
\label{ex_details}

For the regression task, we used summary statistics computed from the \emph{mechanistic signals} extracted at generation step $t=30$, matching the sample length of $30$ tokens. Concretely, for each signal defined in \ref{signals}, we computed a fixed set of descriptive statistics---\emph{mean}, \emph{median}, \emph{range}, \emph{skewness}, \emph{kurtosis}, \emph{variance}, \emph{standard deviation}, \emph{minimum}, and \emph{maximum}---and concatenated them to form the regression feature vector after applying a \emph{Standard Scaler}.

To ensure robustness to stochasticity, we evaluated the pipeline using five random seeds, $\{22, 42, 31, 61, 10\}$. Each seed was used consistently for both the data splitting procedure (via sklearn's \emph{GroupShuffleSplit}) and the regression model initialization. The predictive model was a \emph{Random Forest regressor} with $200$ trees, bootstrap sampling enabled, and $\texttt{max\_features}=\texttt{"sqrt"}$; all other settings followed standard defaults, including unconstrained tree depth and $\texttt{min\_samples\_split}=2$ and $\texttt{min\_samples\_leaf}=1$. Importantly, we performed \emph{no hyperparameter tuning}; all reported results are based on this fixed configuration across seeds.

\section{Similar \ac{nbf} and different \ac{kl}}
\label{similar_different}
In Section~\ref{entropy_kl}, two nearly identical experimental configurations are compared with respect to their \ac{nbf} behavior. Despite their similarity in \ac{nbf}, the configurations exhibit distinct \ac{kl} divergence dynamics and different final steering scores. In both configurations, the model is \emph{Gemma-2-9b}, the target concept is \emph{Christianity}, and the steering vector is extracted using the \ac{caa} method.

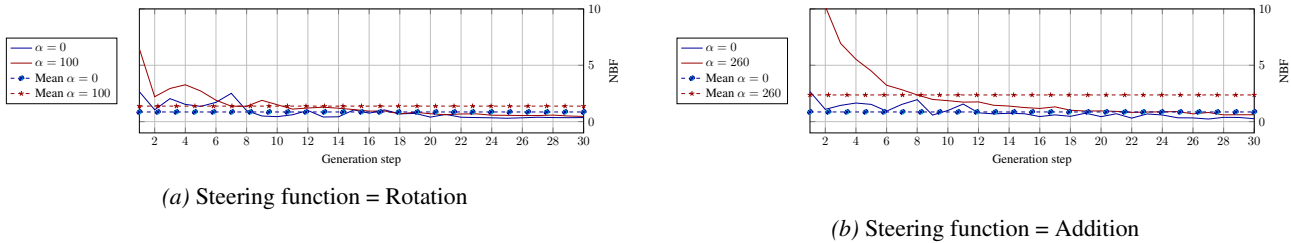
\begin{figure}[ht]
    \centering
    \begin{subfigure}[t]{0.48\textwidth}
        \vspace{0pt}
        \input{graphs/9b_Christian_rotation_act_5}
        \caption{Steering function = Rotation}
        \label{fig:sub1}
    \end{subfigure}
    \hfill
    \begin{subfigure}[t]{0.48\textwidth}
        \vspace{0pt}
        \input{graphs/9b_Christian_addition_act_13}
        \caption{Steering function = Addition}
        \label{fig:sub2}
    \end{subfigure}
    \caption{Similarity betwen the behaviour of the models in terms of \ac{nbf} while the presens of big gap betwen scores $\Delta P(\mathcal{T})>0.1$ probably resulted by different \ac{kl} behavior.}
    \label{fig:side_by_side}
\end{figure}

\section{\ac{llm}-Judge Prompt}
\label{prompt}

To ensure consistent and reproducible evaluation across all experiments, \ac{llm}-based judgments were collected using a standardized zero-shot prompt, following the methodology of \cite{sooInterpretableSteeringLarge2025}. An example prompt for the target concept \emph{angry} is provided below:

\begin{verbatim}
System role: You score texts generated by a language model based on the following
criteria:
score: "The text contains angry words or angry sentiment."
coherence: "The text is coherent and grammatically correct."
Provide integer scores from 1 to 10.
Output JSON only: {"score": int, "coherence": int}
\end{verbatim}

The input to the prompt accompanied with the generated text consists of the generated text to be evaluated. This prompt was applied uniformly across all concepts, steering methods, and base models to ensure comparability of \ac{llm}-judge annotations.

\section{\ac{llm}-Judge Prompt}
\label{regression_overview}

Figure~\ref{fig:distribution} presents the distributions of the ground-truth and predicted values for both \ac{llm} judges. As shown, annotations provided by \gmn exhibit higher variance, which results in greater expressive capacity for the regression model. In contrast, \gpt annotations display lower variance, leading to reduced sensitivity in the learned regression mapping and consequently lower predictive reliability under \gpt supervision.
\input{graphs/regression}
\vspace{11cm}

%% file: graphs/9b_Christian_rotation_act_5.tex
\resizebox{\columnwidth}{!}{%
\begin{tikzpicture}
\begin{axis}[
    width=14cm,
    height=5.05cm,
    grid=both,
    xmin=1,
    xmax=30,
    ymin=-1,
    ymax=10,
    xlabel={Generation step},
    ylabel={\ac{nbf}},
    legend style={
        at={(-0.05,0.5)},
        anchor=east
    },
    yticklabel pos=right,
    ylabel near ticks,
    legend columns=1,
    legend cell align=left,
    cycle list name=exotic,
]

\addplot [color=blue!60!black, thick] coordinates {
(1, 2.66) (2, 1.05) (3, 2.04) (4, 1.53) (5, 1.35) (6, 1.71) (7, 2.51) (8, 0.98) (9, 0.5) (10, 0.44) (11, 0.61) (12, 0.99) (13, 0.41) (14, 0.43) (15, 1.04) (16, 0.75) (17, 1.04) (18, 0.69) (19, 0.73) (20, 0.4) (21, 0.64) (22, 0.39) (23, 0.36) (24, 0.34) (25, 0.3) (26, 0.34) (27, 0.39) (28, 0.36) (29, 0.35) (30, 0.36)
};
\addlegendentry{$\alpha=0$}

\addplot [color=red!60!black, thick] coordinates {
(1, 6.47) (2, 2.21) (3, 2.94) (4, 3.27) (5, 2.72) (6, 1.9) (7, 1.37) (8, 1.36) (9, 1.89) (10, 1.5) (11, 1.11) (12, 1.21) (13, 1.28) (14, 1.19) (15, 1.07) (16, 0.93) (17, 0.95) (18, 0.69) (19, 0.83) (20, 0.7) (21, 0.61) (22, 0.68) (23, 0.69) (24, 0.57) (25, 0.56) (26, 0.54) (27, 0.54) (28, 0.58) (29, 0.5) (30, 0.46)

};
\addlegendentry{$\alpha=100$}

\addplot+[domain=1:30, dashed, blue!60!black] {0.8565540362466572};
\addlegendentry{Mean $\alpha=0$}

\addplot+[domain=1:30, dashed, red!60!black] {1.3766179509482508};
\addlegendentry{Mean $\alpha=100$}

    \end{axis}
    \end{tikzpicture}
    }
  

%% file: graphs/9b_Christian_addition_act_13.tex
\resizebox{\columnwidth}{!}{%
    \begin{tikzpicture}
    \begin{axis}[
        width=14cm, 
        height=5.05cm,
        grid=both,
        xmin=1,
        xmax=30,
        ymin=-1,
        ymax=10,
        xlabel={Generation step},
        ylabel={\ac{nbf}},
        legend style={
            at={(-0.05,0.5)},
            anchor=east
        },
        yticklabel pos=right,
        ylabel near ticks,
        legend columns=1,
        legend cell align=left,
        cycle list name=exotic,
    ]

    \addplot [color=blue!60!black, thick] coordinates {
    (1, 2.66) (2, 1.09) (3, 1.44) (4, 1.65) (5, 1.53) (6, 0.93) (7, 1.53) (8, 1.95) (9, 0.58) (10, 0.99) (11, 1.56) (12, 0.78) (13, 0.69) (14, 0.74) (15, 0.7) (16, 0.45) (17, 0.6) (18, 0.47) (19, 0.75) (20, 0.44) (21, 0.7) (22, 0.31) (23, 0.68) (24, 0.59) (25, 0.34) (26, 0.33) (27, 0.23) (28, 0.38) (29, 0.37) (30, 0.26)
    };
    \addlegendentry{$\alpha=0$}
    
    \addplot [color=red!60!black, thick] coordinates {
    (1, 10.94) (2, 10.23) (3, 6.94) (4, 5.53) (5, 4.49) (6, 3.23) (7, 2.83) (8, 2.37) (9, 1.97) (10, 1.86) (11, 1.73) (12, 1.75) (13, 1.45) (14, 1.38) (15, 1.25) (16, 1.17) (17, 1.31) (18, 1.01) (19, 0.94) (20, 0.95) (21, 0.91) (22, 0.82) (23, 0.85) (24, 0.87) (25, 0.89) (26, 0.69) (27, 0.81) (28, 0.6) (29, 0.61) (30, 0.61)
    };
    \addlegendentry{$\alpha=260$}
    
    \addplot+[domain=1:30, dashed, blue!60!black] {0.8580232335299643};
    \addlegendentry{Mean $\alpha=0$}
    
    \addplot+[domain=1:30, dashed, red!60!black] {2.366531482276573};
    \addlegendentry{Mean $\alpha=260$}

    \end{axis}
    \end{tikzpicture}
    }

%% file: graphs/regression.tex
\pgfplotsset{
  myscatter/.style={
    width=\linewidth,
    height=7cm,
    grid=both,
    axis equal image,
    xlabel={True value},
    ylabel={Predicted value},
    legend style={at={(0.02,0.98)},anchor=north west},
    xmin=0, xmax=.6,
    ymin=0, ymax=.6,
  }
}


\begin{figure}[htbp]
\centering

\begin{subfigure}[t]{0.49\textwidth}
\centering
\begin{tikzpicture}
\begin{axis}[
  myscatter,
]
\addplot[
  only marks,
  mark=*,
  mark size=1pt,
  blue!60!black,
] coordinates {
(0.012698703342013888, 0.014494282286844127)
(0.0231421199845679, 0.04676908987539781)
(0.059305450062692895, 0.07314230930658033)
(0.11604101863908178, 0.12808066850827055)
(0.14869972511574076, 0.11558895063988954)
(0.1633029513888889, 0.1296630435519749)
(0.1784289324725116, 0.11768080252188225)
(0.16411957917390044, 0.11515463840814279)
(0.1517605251736111, 0.12319084450050642)
(0.12327406141493055, 0.13277407610857928)
(0.10638691466531634, 0.102989867410542)
(0.0893727997202932, 0.058755077785915814)
(0.08516664858217592, 0.06414897023895645)
(0.06039617091049382, 0.043109173244900184)
(0.049183786651234566, 0.03621395158179013)
(0.03963517554012346, 0.024799561677155673)
(0.00902416087962963, 0.012592180040147574)
(0.032101478105709874, 0.0704427780339747)
(0.08615620930989583, 0.07531290030773775)
(0.14922153802565585, 0.11502585611225634)
(0.18829157323013118, 0.1297774844699435)
(0.1805872034143518, 0.1384738724320023)
(0.15085177951388887, 0.12280095983434615)
(0.09696225766782406, 0.10013210720486118)
(0.06948061342592592, 0.0968546916820385)
(0.038811577690972224, 0.09694053461522226)
(0.013916392385223766, 0.016688183443045897)
(0.00501806942033179, 0.030551386703679587)
(0.0003729926215277778, 0.03086481447573061)
(0.0, 0.01899202982584636)
(0.0, 0.013806043083285107)
(0.0, 0.018415723259066354)
(0.003617086528260031, 0.025856638778874882)
(0.010662314332561729, 0.08387022607120467)
(0.04691682038483796, 0.09861691321855709)
(0.09070653091242283, 0.12173122641481003)
(0.09030791859567901, 0.12644655298303667)
(0.05864235206886573, 0.04008399680808739)
(0.03488799672067901, 0.0540118389365114)
(0.01226109634211034, 0.06071023069782026)
(0.0028322949821566355, 0.03287157600308644)
(0.00025544343171296296, 0.03207396236466773)
(4.671826774691358e-05, 0.02875497040925203)
(0.0, 0.03357768117645641)
(0.0, 0.025736509723427847)
(0.0, 0.03395940239046825)
(0.0, 0.034419886271158856)
(0.0, 0.03157153565206643)
(0.0008290608723958333, 0.020812772586021894)
(0.018264205367476853, 0.07702465857988519)
(0.09437391493055554, 0.08237486474307962)
(0.24352273823302467, 0.15946784502194236)
(0.3506695782696759, 0.15684229156117374)
(0.33956306363329475, 0.16506573806574307)
(0.3293961889949846, 0.15619378454891245)
(0.2800010398582176, 0.1154225648479697)
(0.24223421826774688, 0.13611707334165213)
(0.16976250542534724, 0.04731436081874522)
(0.10775059829523533, 0.13336378874602137)
(0.07729518560715665, 0.06663386121208288)
(0.0321142879533179, 0.086204801959756)
(0.011661482445987652, 0.03290117899576822)
(0.0033652223186728387, 0.029980441434883782)
(0.0005415928216628086, 0.058875637289918556)
(0.0013365568938078704, 0.01041509651843412)
(0.03286366403838734, 0.06905410860791622)
(0.15508317358699847, 0.10954852351435906)
(0.28643798828125, 0.14996029229811667)
(0.3208258358048804, 0.2210933148419414)
(0.3515794542100695, 0.19730966638635697)
(0.26654655550733025, 0.14210014720021938)
(0.1466640896267361, 0.037997535422996255)
(0.045770339023919755, 0.027718862839687004)
(0.01114456741898148, 0.09638978840392308)
(0.002334971486786265, 0.031370770489728014)
(0.00013563368055555556, 0.018752237013828606)
(1.5070408950617283e-06, 0.025822655006691245)
(0.0, 0.00833693798677421)
(0.0, 0.009012164834104943)
(0.0, 0.022520209418402763)
(0.0, 0.02258151772581502)
(0.002453274197048611, 0.04009996202256947)
(0.021270751953125, 0.06937552463861159)
(0.06310206283757716, 0.09484657664357896)
(0.12805627893518517, 0.12615621025179638)
(0.17692170319733797, 0.12250642093611351)
(0.17651141131365738, 0.1247235908037351)
(0.1277953724802276, 0.08431464207025226)
(0.06179319782021605, 0.06442904342839746)
(0.012754087094907406, 0.043714335877218374)
(0.0010579427083333333, 0.03154968826859088)
(6.273057725694444e-05, 0.02153348286946615)
(0.0, 0.029521879031334394)
(0.0, 0.044145364643615)
(0.0, 0.04178516058274256)
(0.0, 0.006598497555579672)
(0.0, 0.014000838479877993)
(0.008578076774691358, 0.06293536904417443)
(0.033672944999035496, 0.04986497196150416)
(0.09066471052758486, 0.11760106404622402)
(0.144280704451196, 0.11395034695849007)
(0.2496612925588349, 0.1183586026415413)
(0.24672067901234568, 0.12767316465024586)
(0.20806206597222224, 0.14146160832157828)
(0.1503827130353009, 0.14371209415388692)
(0.08886341989776234, 0.10013939845709151)
(0.07035997178819445, 0.04678921734845196)
(0.04333646797839506, 0.025161624484592)
(0.029628423996913584, 0.009366483335141782)
(0.040481002242476846, 0.007169229954849057)
(0.011029278790509259, 0.024608628072856395)
(0.0012427435980902778, 0.015750621277608977)
(0.01440712257667824, 0.022083886228961715)
(0.053320237147955245, 0.06314365810818141)
(0.10294652868200231, 0.07706015127676503)
(0.0854665497202932, 0.12278765643084487)
(0.03795068057966821, 0.09656467579029224)
(0.008076608916859566, 0.07321517850145884)
(0.0018856849199459875, 0.05260489569769967)
(0.00032778139467592586, 0.05990780300564239)
(3.164785879629629e-05, 0.044154355084454554)
(9.795765817901234e-06, 0.03803779036910443)
(0.0, 0.05114595766420718)
(0.0, 0.034910327299141584)
(0.0, 0.023966107780550718)
(1.5070408950617283e-06, 0.028188856148425444)
(3.0140817901234566e-06, 0.041178077415183735)
(5.651403356481481e-07, 0.03476846671398774)
(0.018817666136188273, 0.009407603887864107)
(0.09161376953125, 0.08371482849121094)
(0.17545572916666666, 0.08149044554910544)
(0.19209006980613427, 0.14615800928186481)
(0.14419404959972992, 0.1148874701982662)
(0.0446296974464699, 0.1315279520293813)
(0.008653428819444444, 0.03711658430688178)
(0.0013258192274305553, 0.042931721063307754)
(0.0002486617476851852, 0.03222429157775123)
(3.202461902006173e-05, 0.028921357143072413)
(7.1584442515432095e-06, 0.03148059986255789)
(1.4693648726851851e-05, 0.02494192523720822)
(4.897882908950617e-06, 0.017841581179771887)
(0.0, 0.014436881924852906)
(1.318660783179012e-06, 0.0169628256338614)
(0.0, 0.017569882428204564)
(0.054494222005208336, 0.009560072157118063)
(0.08701277367862653, 0.09620960282690724)
(0.1129116482204861, 0.101236604290244)
(0.12004993580005785, 0.12151543793854888)
(0.13017216435185186, 0.12077596782166275)
(0.13287805627893517, 0.13327207871425295)
(0.09200032552083334, 0.13618826218593263)
(0.0827374870394483, 0.1551731580569419)
(0.0523048683449074, 0.14885823756088437)
(0.04393627025462962, 0.12773550057116842)
(0.032830132378472224, 0.09867862089180647)
(0.0207312313126929, 0.09513969609766823)
(0.013361612955729166, 0.09550850762261279)
(0.011730617947048608, 0.09307422025704087)
(0.007897647810570988, 0.11168823713137772)
(0.0038203486689814808, 0.05443567817593796)
(0.003826753592785494, 0.04090242550696854)
(0.015096028645833332, 0.06421693684142311)
(0.11383056640625, 0.07447438746322822)
(0.2144195179880401, 0.13056696479703178)
(0.2513864776234568, 0.10994680616590713)
(0.21575136537905093, 0.11163555003978588)
(0.21946132330246912, 0.1150740315001689)
(0.15620478877314814, 0.09735814883385178)
(0.12286150896990741, 0.09758141694245519)
(0.08464860327449845, 0.09686277695644048)
(0.0457970889998071, 0.056100178471317964)
(0.032510074568383486, 0.021871504607024018)
(0.008587872540509259, 0.021553663088951572)
(0.0014173719618055555, 0.020223874692563655)
(7.252634307484568e-05, 0.028568673781406723)
(0.0, 0.047734590695228146)
(0.004701214072145062, 0.016486464135440748)
(0.007344563802083332, 0.10471395704481339)
(0.021409776475694444, 0.10496205176836179)
(0.0721610740379051, 0.11616192853009259)
(0.11441454475308642, 0.12539969173478496)
(0.18715036651234565, 0.1408848590615354)
(0.23661973741319445, 0.12919913680465134)
(0.23198615180121526, 0.13203286630135985)
(0.21584875789689426, 0.13348459973747354)
(0.17668622805748455, 0.11728985350808989)
(0.0968288845486111, 0.07464005929452401)
(0.04178383909625771, 0.0637886565408589)
(0.007300106095679011, 0.03924750339837721)
(0.0006499113859953703, 0.025544263109748732)
(7.911964699074072e-06, 0.029793371506679208)
(1.8838011188271604e-07, 0.03106886969672308)
(0.011064317491319444, 0.04683553907606337)
(0.024435161072530864, 0.01562573656623746)
(0.04413972077546296, 0.03125226903844764)
(0.10592820909288193, 0.11246862152476368)
(0.1106426097728588, 0.10568115987895446)
(0.08887660650559415, 0.14019881448627994)
(0.08533430688175153, 0.1548050133975935)
(0.0734339584538966, 0.10877396760163485)
(0.06596600567853009, 0.12251626756456163)
(0.03306221667631173, 0.13014242289978792)
(0.010439649040316357, 0.06493034362792974)
(0.0009562174479166667, 0.05288731233573256)
(0.0009036593967013888, 0.014062419938452437)
(1.6389069733796295e-05, 0.010863830189646023)
(0.0, 0.008602662263093171)
(0.0, 0.010429689383801117)
(0.01088799370659722, 0.019112756988148613)
(0.014546712239583334, 0.04485593630943775)
(0.03223183714313271, 0.07068147447374139)
(0.06831171483169367, 0.17339428936993623)
(0.1372612847222222, 0.19070358747317456)
(0.1802345558449074, 0.19853045616620868)
(0.21376809956114967, 0.1878736472424166)
(0.22760913990162035, 0.17110260480715891)
(0.22855631510416666, 0.1555177900526257)
(0.1987033420138889, 0.13704586641288097)
(0.1846539532696759, 0.13058375323260268)
(0.16114524558738424, 0.1706343266993393)
(0.14072747878086417, 0.11859055884090466)
(0.13124894507137344, 0.12384668138292099)
(0.10478606047453702, 0.12246623003924344)
(0.09266266999421295, 0.1153044260284047)
(0.0019437059944058639, 0.0030452257321204657)
(0.021408646195023147, 0.0723500013940128)
(0.04427158685378087, 0.04144592002586076)
(0.0951189582730517, 0.09826901188603154)
(0.154926441333912, 0.13316305184069963)
(0.2405478395061728, 0.2214945965048706)
(0.2483181423611111, 0.1366238488091364)
(0.19749431845582557, 0.200722957658179)
(0.1553802490234375, 0.17900854605215563)
(0.08249541859567901, 0.12387588406786502)
(0.03895493495611496, 0.11375792515130682)
(0.012134316526813273, 0.043854059290002904)
(0.0004362883391203703, 0.02576891675407506)
(0.0, 0.006395743099259743)
(0.0, 0.012638723114390436)
(0.0, 0.006256029105480808)
(0.0, 0.006392791182906539)
(0.03525175871672454, 0.04401035638503086)
(0.3107198079427083, 0.1849056149706428)
(0.30106495044849535, 0.2053737998303071)
(0.2876262900270062, 0.1700872435393156)
(0.27651242856626157, 0.20360717585057372)
(0.25031082718460645, 0.09736107720269098)
(0.2305668960382909, 0.18381569756401905)
(0.18514939296392743, 0.1427866608419536)
(0.1468799732349537, 0.1331877494152681)
(0.11163254726080246, 0.06893875781400707)
(0.08135082103587962, 0.1234832631805797)
(0.04466869212962962, 0.08092135582441166)
(0.02887339650848765, 0.05567768426589023)
(0.012489601417824072, 0.06377405896598912)
(0.004148130063657407, 0.06951048297646607)
(0.002532958984375, 0.02412541989926937)
(0.0, 0.03249350182804063)
(0.005598656925154322, 0.044113483193479935)
(0.01368091724537037, 0.09990182382089122)
(0.042529447579089504, 0.13721672717435862)
(0.10457036524643132, 0.16137643037018962)
(0.10660166799286265, 0.19521257376965193)
(0.10185750325520834, 0.1140085196789399)
(0.05711270556037808, 0.10099559077510124)
(0.025845751350308643, 0.09938265294204514)
(0.008527779284818673, 0.09435981185347946)
(0.001739501953125, 0.1278740164674358)
(0.0, 0.07423423296139567)
(0.0002394311222029321, 0.09838061297381373)
(0.0, 0.08238135349603348)
(0.0, 0.06533148494767556)
(0.013679786964699073, 0.027304683732397746)
(0.030033817997685185, 0.06577424037603687)
(0.06069550690827546, 0.03101665355541088)
(0.0788464958285108, 0.12333220399456256)
(0.11376425660686729, 0.21223238061975555)
(0.1037575050636574, 0.1636490913673684)
(0.07836405436197916, 0.16318058343581213)
(0.04112337842399691, 0.12975476206084818)
(0.016574435763888888, 0.13315326361008606)
(0.006449758270640431, 0.08261222274215135)
(0.0013687698929398147, 0.08937066725742673)
(0.001475393036265432, 0.10986174406828698)
(0.0005321738160686729, 0.10017435709635412)
(0.00033569335937499995, 0.09486905510042921)
(0.00016389069733796293, 0.06484252081976999)
(5.6514033564814815e-05, 0.06923355950249566)
(0.014912923177083332, 0.00512457576798804)
(0.035786193094135804, 0.056326576515480335)
(0.0568738395785108, 0.05849411575882528)
(0.11180623372395833, 0.1307168692129631)
(0.13606054988908178, 0.11471175299750434)
(0.24829798568913963, 0.1346272701687282)
(0.29908695927372686, 0.1331673987117814)
(0.3599478639202353, 0.153636951211058)
(0.33305603780864196, 0.12988331170729636)
(0.3078782823350695, 0.142985247388298)
(0.2737856264467592, 0.10525210816183199)
(0.1714552891107253, 0.10284580183617864)
(0.10783329716435183, 0.0945480120623553)
(0.08070335859133872, 0.08729287700888559)
(0.0649052372685185, 0.09110219884801796)
(0.05346510145399305, 0.07631218239113143)
(0.0672008373119213, 0.02419491803204573)
(0.04659393687307098, 0.03126007927788633)
(0.07116623866705248, 0.04582514256606868)
(0.10412748360339506, 0.1249469248453776)
(0.14965217496141975, 0.1295466604350525)
(0.1618072133005401, 0.19681709854691107)
(0.1510117142288773, 0.23986030861183447)
(0.15746768904320985, 0.16352291719413078)
(0.14377283166956017, 0.13233269962263677)
(0.09449146412037036, 0.12278955906997475)
(0.033854920187114196, 0.09291544407974049)
(0.002893518518518518, 0.028394184818974266)
(0.0, 0.008202681364836522)
(0.0, 0.012576055644470976)
(0.0, 0.011019501862702553)
(0.0, 0.028303313078703696)
(0.0013996642312885804, 0.00632137722439236)
(0.005035400390625, 0.06346996731228302)
(0.013661702473958334, 0.07584933999144)
(0.07768230673707562, 0.11685303770465616)
(0.11492693865740738, 0.10065770090362172)
(0.16733466254340276, 0.16971487163025656)
(0.15239197530864196, 0.19909964290665993)
(0.1251000298394097, 0.26559990400149486)
(0.08674150631751543, 0.07642251379695941)
(0.018110487196180552, 0.06512025244442039)
(7.04541618441358e-05, 0.0438937114197531)
(0.0, 0.036255528956283756)
(0.0, 0.03490591449502075)
(0.0, 0.022181608412000852)
(0.0, 0.019331760171018973)
(0.0, 0.019193701096522943)
(0.004487967785493826, 0.012564574818552275)
(0.020004084080825615, 0.03451215579185962)
(0.060317051263503084, 0.030444679731204186)
(0.201361385392554, 0.08466103918758439)
(0.28967548888406636, 0.13625238395031586)
(0.3347085081500772, 0.14215653690291044)
(0.3200276692708333, 0.19219458968551073)
(0.20881897726176699, 0.0846075128625941)
(0.1375536506558642, 0.12475886968918777)
(0.059531882957175916, 0.09164308241855951)
(0.03625657823350694, 0.06806278181664743)
(0.013783772786458334, 0.03884327547049817)
(0.003145571108217592, 0.019799197161639172)
(6.329571759259258e-05, 0.013463610425407503)
(5.274643132716048e-06, 0.011361653834213448)
(0.0, 0.010229641007788384)
};
\addlegendentry{Samples}

\addplot[
  domain=0:1,
  samples=2,
  thick,
  dashed,
  black
] {x};
\addlegendentry{$y=x$}
\end{axis}
\end{tikzpicture}

\caption{Judge LLM $=$ \emph{ChatGPT-4o-mini}}
\label{fig:panel-a}
\end{subfigure}
\hfill
\begin{subfigure}[t]{0.49\textwidth}
\centering
\begin{tikzpicture}
\begin{axis}[
  myscatter,
  ylabel={}, 
]
\addplot[
  only marks,
  mark=square*,
  mark size=1pt,
  red!60!black,
] coordinates {
(0.04726833767361111, 0.046210298024785275)
(0.13611009385850695, 0.1135318083283485)
(0.20431518554687503, 0.1960931553190265)
(0.3127856807210902, 0.24284203474373317)
(0.4128613884066358, 0.24033831811832737)
(0.409041536730887, 0.22178012668232394)
(0.3739053231698496, 0.2139825788842537)
(0.3604323628400343, 0.2018128815987172)
(0.30644905424446867, 0.2317432840097302)
(0.23885957694347995, 0.2523779792103646)
(0.18600037200074399, 0.1377464071375607)
(0.15132034941244318, 0.06964036267597927)
(0.09881422924901184, 0.05522545433130079)
(0.07779006298753337, 0.027972231421776933)
(0.06277927292921526, 0.021889781302492643)
(0.03219720019686952, 0.021042256220282617)
(0.040295259452160496, 0.018115142672590005)
(0.14024522569444445, 0.16011308327730622)
(0.24113802383698577, 0.21176150120841497)
(0.4039357503255208, 0.26230132419608737)
(0.4291289674911358, 0.28568781164529444)
(0.42160384958738856, 0.2648333187465418)
(0.30050153055251294, 0.20696594690216805)
(0.20192703192443937, 0.1859189018422066)
(0.09028560623578834, 0.11490596187620401)
(0.021604584144537267, 0.14775681197875876)
(0.005059370075784468, 0.03482954283992088)
(0.00047220683333881427, 0.03981652178841137)
(0.0005494028507026589, 0.029794955707285898)
(0.0001148150444449037, 0.022005893418494685)
(0.0, 0.008325198722245209)
(0.00027019807126034, 0.00974977098136577)
(0.008707682291666668, 0.05597599536875962)
(0.0345850815007716, 0.17181537843078498)
(0.17110114343486124, 0.20336235472637096)
(0.2703386471595293, 0.2048163201898966)
(0.3244063765914352, 0.1819302321769037)
(0.2459301028569259, 0.08950856090867604)
(0.1315697564019097, 0.1053533503266158)
(0.08136268898292823, 0.09375483048894978)
(0.01770245587384259, 0.07232003323445821)
(0.002963595920138889, 0.0650860346648828)
(0.00012169355227623457, 0.07650366863394661)
(2.637321566358024e-06, 0.08206951192315659)
(5.651403356481481e-07, 0.060721885404810855)
(1.8838011188271604e-07, 0.03818763237459597)
(0.0, 0.02418368331947784)
(0.0, 0.04114585507108009)
(0.0026884246799663944, 0.05395516113259233)
(0.05542689193913966, 0.17100514660514676)
(0.2934133270640432, 0.19739154720067845)
(0.5074909351490162, 0.2961602661466052)
(0.5737869827835648, 0.3090346953565675)
(0.53169627248505, 0.2705757519849288)
(0.460205078125, 0.29168805559239736)
(0.39547559950086797, 0.1416417691692033)
(0.2874077690972222, 0.1697798002610757)
(0.206695556640625, 0.07786444559680634)
(0.1360198597849151, 0.18548582835508365)
(0.11582965615354938, 0.046520647519130966)
(0.07847877785011573, 0.08311144100547735)
(0.029947916666666664, 0.03744569505185841)
(0.011535644531249998, 0.0318429725362447)
(0.0034970883969907404, 0.07039138319188323)
(0.00033267927758487653, 0.041150329950684704)
(0.08434870213638117, 0.15737445280163515)
(0.36162049093364196, 0.2456708408026736)
(0.5824822319878472, 0.2820322239738339)
(0.5715301890432098, 0.3149248717824843)
(0.481970516251929, 0.3379837925308824)
(0.35772422507957174, 0.25428764575763885)
(0.21872664086612653, 0.06282969649521189)
(0.06515898527922453, 0.04073054464815154)
(0.016520935812114196, 0.08329756188507861)
(0.002017550998263889, 0.05259813200426954)
(0.00033286765769675923, 0.03657643728715089)
(0.0, 0.020537431982278878)
(8.288724922839505e-05, 0.009660793074117708)
(0.00033795392071759255, 0.009784955254649256)
(0.00015371817129629627, 0.01932597271302738)
(0.008453745900848766, 0.06094068700567965)
(0.015202086648823302, 0.11247191931580916)
(0.03334139600212191, 0.1463130513655704)
(0.08750632957175924, 0.16898530221715174)
(0.1691446186583719, 0.22173158793694384)
(0.20440297067901234, 0.24136958866539002)
(0.1408962673611111, 0.1981389344846256)
(0.09429253472222221, 0.10854741762214051)
(0.02538572711709105, 0.0811249445207027)
(0.0010505958839699073, 0.06946488808839174)
(0.0, 0.03629346440216328)
(0.0, 0.016992574368739674)
(0.0, 0.05596240082277428)
(0.0, 0.07780573429021458)
(0.0, 0.06338536391180492)
(0.0, 0.00981123485162587)
(0.007153169608410493, 0.036090040673515705)
(0.026153564453124997, 0.1286003892645928)
(0.08775687512056328, 0.11817964873379198)
(0.19235907660590276, 0.23793152257880806)
(0.3204639576099537, 0.2656554559196412)
(0.33994547526041663, 0.2451418992806017)
(0.3243148238570602, 0.2352517359930275)
(0.22525156280140818, 0.24918795259400292)
(0.09177935564959491, 0.20151518354976375)
(0.014083297164351851, 0.13045623275911322)
(0.004202383535879629, 0.051945424776038675)
(0.0003712972005208333, 0.03546207941617423)
(0.0, 0.013677904400776222)
(0.0, 0.014326352463693072)
(0.0, 0.01551490096618947)
(0.0, 0.005172195390506316)
(0.05606907974054784, 0.06353837899090158)
(0.13363402566792054, 0.17921533607424492)
(0.24278654875578703, 0.1938613955308225)
(0.19196875301408178, 0.25877425715798896)
(0.04326695571711033, 0.1410437993103086)
(0.006639457043306327, 0.15565275675557108)
(0.0007165979456018518, 0.09979084376294863)
(0.00023133077739197528, 0.08357174083771221)
(0.0, 0.05584838604324457)
(0.0, 0.04494028471903086)
(0.0, 0.059558060514751425)
(0.0, 0.032046118446857196)
(0.0, 0.030489104138812794)
(0.0, 0.01818895412395836)
(0.0, 0.032363535266696816)
(0.0, 0.016915138150438838)
(0.05585093557098765, 0.03809985101627503)
(0.2346123589409722, 0.2058402051961157)
(0.43402641908857464, 0.20912102801743604)
(0.40595367808400845, 0.28968188947168017)
(0.17908016251929013, 0.19117543546452198)
(0.03315647827758555, 0.2431239027111828)
(0.004648844401041667, 0.0598790824457935)
(0.0, 0.04871219458118598)
(0.0001314893180941358, 0.040926503399821124)
(0.0, 0.031107837361945098)
(0.0, 0.032325213358923625)
(0.0, 0.010900336666823189)
(0.0, 0.005212514429916568)
(0.0, 0.009882133389696656)
(0.0, 0.00883629787115403)
(0.00024264171899696695, 0.012830415890540608)
(0.1580305770335246, 0.031393479305890926)
(0.2791148998119213, 0.19675960340953122)
(0.3071537724247685, 0.220008761621319)
(0.28017246576003085, 0.2746124885254857)
(0.30527279700761956, 0.2708518534457316)
(0.23903401692708331, 0.27905217104034674)
(0.215972900390625, 0.22677025386481073)
(0.17796871985918208, 0.32577898186035925)
(0.11360677083333334, 0.2656003953829558)
(0.08861400462962962, 0.233302535189967)
(0.070770263671875, 0.19901027644337746)
(0.04125336070119598, 0.20666981653941766)
(0.031958685980902776, 0.20588655524636176)
(0.018627025462962965, 0.16616192652626713)
(0.012010362413194444, 0.22818885963072702)
(0.006227846498842592, 0.08302831739525171)
(0.0, 0.07395961519205907)
(0.03253908510561342, 0.11724428429555125)
(0.2615047501929012, 0.1454550335744144)
(0.4255265600887346, 0.23297108749859027)
(0.4191020447530864, 0.220215209386599)
(0.40679856288580246, 0.2022151712645337)
(0.3436733292944637, 0.18371033313459395)
(0.2603313304759838, 0.1519936699663654)
(0.1580810546875, 0.12478420050510296)
(0.1070240162037037, 0.1764376623054073)
(0.0914775707103588, 0.07253541813537977)
(0.0441523422429591, 0.029345262444311702)
(0.014880710177951388, 0.04096264188598824)
(0.0023820665147569445, 0.014692595443947562)
(0.00012640305507330245, 0.03825540422961616)
(2.82570167824074e-06, 0.05520742399069479)
(0.0006171332465277778, 0.05662733760880833)
(0.011653947241512346, 0.2205939582501071)
(0.045220834237557864, 0.20875012540302534)
(0.13058151433497298, 0.27355453674809466)
(0.19550182201244212, 0.22781561538428505)
(0.36063300238715273, 0.2761372159025988)
(0.39928784782503857, 0.2785250356858563)
(0.449544835973669, 0.2889917624595253)
(0.3973616611810378, 0.2322440661830463)
(0.3028880931712963, 0.18502666665015638)
(0.13811841423128857, 0.08912573271279578)
(0.04397658359857253, 0.06570609127907287)
(0.0066375732421875, 0.06919694632832953)
(0.0, 0.035584073459759405)
(0.0, 0.04840189708172117)
(0.0, 0.05639518233615072)
(0.04834022051022376, 0.10133655910934929)
(0.07585972915461034, 0.058522634367256285)
(0.2247585720486111, 0.0836912511163053)
(0.315100399064429, 0.2536418324123541)
(0.31714771412037035, 0.2407448461648666)
(0.2589797031732253, 0.24540121420000585)
(0.25461606626157407, 0.21373268283258764)
(0.25518064145688657, 0.1741241847925185)
(0.14603226273148148, 0.12933910899507967)
(0.07952202690972221, 0.15449335637822798)
(0.015979342990451388, 0.06843417855969626)
(0.0020966706452546294, 0.052016014391599155)
(0.0005086263020833333, 0.0084995702609008)
(0.0, 0.012564573876651714)
(0.0, 0.009015245790834774)
(0.0, 0.003993861778370416)
(0.04346814567660108, 0.05805612917299631)
(0.08540834026572144, 0.11720287118624097)
(0.1401451958550347, 0.15556898152867427)
(0.2164144633728781, 0.3510382175952666)
(0.31707160855516975, 0.3992469640828763)
(0.4522443229769483, 0.39455253811991864)
(0.528045654296875, 0.3723507210502636)
(0.5901493284468051, 0.360459572571323)
(0.5608868383954889, 0.3450622379605847)
(0.5401811011043595, 0.29286997006762994)
(0.46800328458664997, 0.2365838402920116)
(0.4181866264967098, 0.30862745881904735)
(0.38843677662037035, 0.21170733469895642)
(0.36191360397947603, 0.2109483443066567)
(0.2986005618248457, 0.18848885134039425)
(0.20169105058834877, 0.15460589905029956)
(0.04321345576533565, 0.015534120206302227)
(0.09597100151909722, 0.1920662409738739)
(0.17716923466435183, 0.12893525852126572)
(0.3176342999493634, 0.24313899932353927)
(0.464353947226413, 0.33409978895705805)
(0.529592443395544, 0.42589927488065454)
(0.6121590696735146, 0.2677196652279488)
(0.5519590024594907, 0.4209451758690709)
(0.3838324343315353, 0.35469645183434545)
(0.22327752760898917, 0.19760253697696334)
(0.07277732685909027, 0.1152805363940462)
(0.015193206791310084, 0.06641726852465443)
(0.001869954357192665, 0.038128090761027474)
(0.001011431841471367, 0.019549034463943467)
(0.000518319164993521, 0.022616813063058527)
(0.0009053548177083333, 0.004078597871539572)
(0.0, 0.01694463744290374)
(0.035912972909432864, 0.08705514761130377)
(0.37855548623167434, 0.24925471967681506)
(0.3739842544367284, 0.2810639109729479)
(0.3497374734760802, 0.2802071793396401)
(0.34092655888310186, 0.2723311569490685)
(0.3128921924615888, 0.15066933653043743)
(0.25756247597078047, 0.2857470073266646)
(0.21032865547839502, 0.1902291696281238)
(0.18390721450617284, 0.1861905482451499)
(0.1429204116632909, 0.13945496097958526)
(0.10268543384693286, 0.1977092346019467)
(0.06372974537037036, 0.11559847698993245)
(0.03876467104311342, 0.09673756720846154)
(0.019493573977623455, 0.09992848872219065)
(0.011716677818769288, 0.10886275424704049)
(0.0, 0.06907504536114885)
(0.012383355034722222, 0.08625694621307364)
(0.01775181146315586, 0.0953476606251986)
(0.024685329861111112, 0.20072839968946204)
(0.09630650649836034, 0.2632919788130625)
(0.21975237057532795, 0.3111044825987566)
(0.18736191737799, 0.3334276133866167)
(0.09239366319444443, 0.2092634667154276)
(0.04018392680603781, 0.15222724065451135)
(0.005614669234664351, 0.20140786056789783)
(0.0007478690441743826, 0.19053338171983342)
(0.0003871211299189815, 0.23405640559697513)
(0.0, 0.1382606724897846)
(0.0, 0.17925641567023387)
(0.0, 0.13082634230039403)
(0.0, 0.10024024063699821)
(0.06713302047164352, 0.07492273454814752)
(0.09446810498649691, 0.1278659162341466)
(0.15470528308256173, 0.12458998600320272)
(0.22761479130497683, 0.25562156322052887)
(0.23687367380401234, 0.40299035440835945)
(0.194538257740162, 0.2842378379995783)
(0.0957183837890625, 0.2957635415402831)
(0.036854685088734566, 0.24490266852186124)
(0.008899830005787035, 0.2828038187979987)
(0.0011076750578703702, 0.1436588813951324)
(0.0, 0.18575889650240623)
(0.0, 0.22249378117396085)
(0.0, 0.21447964479249776)
(0.0, 0.19289965629616213)
(0.0, 0.11145279636471683)
(0.0, 0.13435249587629053)
(0.07663001543209877, 0.02257316701988091)
(0.10808685679494599, 0.13523216842831065)
(0.16101677035108025, 0.14519631408875086)
(0.28628238630883485, 0.2764451669345019)
(0.390790436726615, 0.27616435473264084)
(0.5208696906949266, 0.2915467272036711)
(0.5143291332103588, 0.2735641228414163)
(0.47133156105324076, 0.29080299027060197)
(0.3952544412495177, 0.27085591304787604)
(0.3670270001446759, 0.29343274246453316)
(0.26223057876398537, 0.2255233237047095)
(0.16546743887442125, 0.20245313096208206)
(0.11543142059702932, 0.2108686553430571)
(0.08179784704137731, 0.18918416791868975)
(0.046295166015625, 0.19200311491508046)
(0.029481299129533175, 0.1669540940437942)
(0.17245370370370366, 0.06811920203090524)
(0.18855134940441742, 0.09484711968237171)
(0.22418099862557866, 0.14355486289354483)
(0.22723162615740738, 0.2648102947497668)
(0.2504323323567708, 0.27366788929297114)
(0.2889197078751929, 0.34457450596746975)
(0.28929722161940585, 0.4085270070259309)
(0.26949470425829475, 0.26582076111365993)
(0.2454136822969323, 0.2888953010554711)
(0.14989179446373455, 0.18987105072223748)
(0.05101634837962962, 0.1278236000602167)
(0.0036282009548611106, 0.03300901169367959)
(1.770773051697531e-05, 0.009144399180114827)
(0.0, 0.017105625230934757)
(0.0, 0.00968496178778163)
(0.0, 0.05293127062636926)
(0.0028211805555555555, 0.02290249481863871)
(0.008097895869502315, 0.14899213478636308)
(0.02377507716049383, 0.14080161056571072)
(0.09157816569010417, 0.20660929349962245)
(0.3064507378472222, 0.21852581222809178)
(0.3323974609375, 0.33699008078818815)
(0.39035297911844136, 0.2571114893592343)
(0.3136800130208333, 0.47575407143712456)
(0.20955139913676696, 0.12555263304189762)
(0.03209752212336034, 0.08687654221016555)
(1.1679566936728394e-05, 0.040970314512169044)
(0.0, 0.033677173039421145)
(0.0, 0.03195730612396028)
(0.0, 0.015929208763704465)
(0.0, 0.01814969412129513)
(1.1302806712962962e-06, 0.01870570053288967)
(0.00520531925154321, 0.02619866714442412)
(0.024702472451292437, 0.08610303973365696)
(0.12616550775221835, 0.10014797422971883)
(0.3942298418209877, 0.1920220644805964)
(0.5406505443431714, 0.26483561344278106)
(0.6128359194155092, 0.3091181160982001)
(0.5657205463927468, 0.3344831247353494)
(0.4197031656901042, 0.10713739592946882)
(0.27069769965277773, 0.1733889910221107)
(0.14383631576726466, 0.12192576022609877)
(0.06920331790123457, 0.10495267234804162)
(0.010091145833333332, 0.05416819968667837)
(0.0015475426191165122, 0.01720511354046103)
(0.00033286765769675923, 0.0042113251734363635)
(0.0, 0.010286561000494313)
(0.0, 0.010126917904147714)
};
\addlegendentry{Samples}

\addplot[
  domain=0:1,
  samples=2,
  thick,
  dashed,
  black
] {x};
\addlegendentry{$y=x$}
\end{axis}
\end{tikzpicture}

\caption{Judge LLM $=$ \emph{Gemini-Flash-2.5}}
\label{fig:panel-b}
\end{subfigure}

\caption{The distribution of test-set predicted values and corresponding ground-truth annotations for seed = 10, evaluated across different \ac{llm} judges.
}
\label{fig:distribution}
\end{figure}
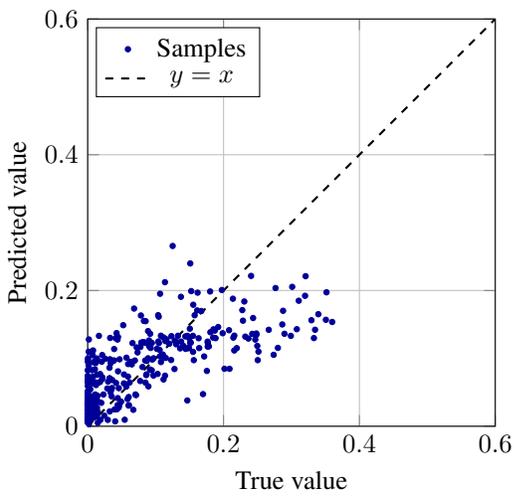
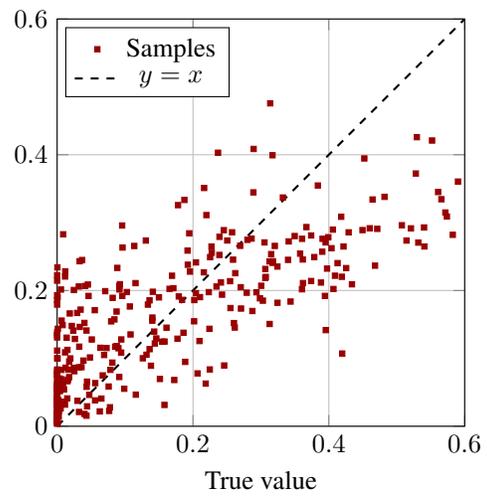

%% file: example_paper.bib
@misc{turnerSteeringLanguageModels2024a,
  title = {Steering {{Language Models With Activation Engineering}}},
  author = {Turner, Alexander Matt and Thiergart, Lisa and Leech, Gavin and Udell, David and Vazquez, Juan J. and Mini, Ulisse and MacDiarmid, Monte},
  year = 2024,
  month = oct,
  number = {arXiv:2308.10248},
  eprint = {2308.10248},
  primaryclass = {cs},
  publisher = {arXiv},
  doi = {10.48550/arXiv.2308.10248},
  urldate = {2026-01-07},
  archiveprefix = {arXiv}
}

@misc{chenDesigningDashboardTransparency2024,
  title = {Designing a {{Dashboard}} for {{Transparency}} and {{Control}} of {{Conversational AI}}},
  author = {Chen, Yida and Wu, Aoyu and DePodesta, Trevor and Yeh, Catherine and Li, Kenneth and Marin, Nicholas Castillo and Patel, Oam and Riecke, Jan and Raval, Shivam and Seow, Olivia and Wattenberg, Martin and Vi{\'e}gas, Fernanda},
  year = 2024,
  month = oct,
  number = {arXiv:2406.07882},
  eprint = {2406.07882},
  primaryclass = {cs},
  publisher = {arXiv},
  urldate = {2024-10-21},
  archiveprefix = {arXiv}
}

@misc{gurneeLanguageModelsRepresent2024,
  title = {Language {{Models Represent Space}} and {{Time}}},
  author = {Gurnee, Wes and Tegmark, Max},
  year = 2024,
  month = mar,
  number = {arXiv:2310.02207},
  eprint = {2310.02207},
  primaryclass = {cs},
  publisher = {arXiv},
  urldate = {2024-10-28},
  archiveprefix = {arXiv}
}

@misc{chenSTEERBENCHBenchmarkEvaluating2025,
  title = {{{STEER-BENCH}}: {{A Benchmark}} for {{Evaluating}} the {{Steerability}} of {{Large Language Models}}},
  shorttitle = {{{STEER-BENCH}}},
  author = {Chen, Kai and He, Zihao and Shi, Taiwei and Lerman, Kristina},
  year = 2025,
  month = jun,
  number = {arXiv:2505.20645},
  eprint = {2505.20645},
  primaryclass = {cs},
  publisher = {arXiv},
  doi = {10.48550/arXiv.2505.20645},
  urldate = {2025-07-22},
  archiveprefix = {arXiv}
}

@misc{jafariEnhancingConversationalAgents2025,
  title = {Enhancing {{Conversational Agents}} with {{Theory}} of {{Mind}}: {{Aligning Beliefs}}, {{Desires}}, and {{Intentions}} for {{Human-Like Interaction}}},
  shorttitle = {Enhancing {{Conversational Agents}} with {{Theory}} of {{Mind}}},
  author = {Jafari, Mehdi and Hua, Devin Yuncheng and Xue, Hao and Salim, Flora},
  year = 2025,
  month = may,
  number = {arXiv:2502.14171},
  eprint = {2502.14171},
  primaryclass = {cs},
  publisher = {arXiv},
  doi = {10.48550/arXiv.2502.14171},
  urldate = {2026-01-07},
  archiveprefix = {arXiv}
}

@misc{zhaoAnalysingResidualStream2024,
  title = {Analysing the {{Residual Stream}} of {{Language Models Under Knowledge Conflicts}}},
  author = {Zhao, Yu and Du, Xiaotang and Hong, Giwon and Gema, Aryo Pradipta and Devoto, Alessio and Wang, Hongru and He, Xuanli and Wong, Kam-Fai and Minervini, Pasquale},
  year = 2024,
  month = oct,
  number = {arXiv:2410.16090},
  eprint = {2410.16090},
  primaryclass = {cs},
  publisher = {arXiv},
  urldate = {2024-11-06},
  archiveprefix = {arXiv}
}

@misc{sharkeyOpenProblemsMechanistic2025,
  title = {Open {{Problems}} in {{Mechanistic Interpretability}}},
  author = {Sharkey, Lee and Chughtai, Bilal and Batson, Joshua and Lindsey, Jack and Wu, Jeff and Bushnaq, Lucius and {Goldowsky-Dill}, Nicholas and Heimersheim, Stefan and Ortega, Alejandro and Bloom, Joseph and Biderman, Stella and {Garriga-Alonso}, Adria and Conmy, Arthur and Nanda, Neel and Rumbelow, Jessica and Wattenberg, Martin and Schoots, Nandi and Miller, Joseph and Michaud, Eric J. and Casper, Stephen and Tegmark, Max and Saunders, William and Bau, David and Todd, Eric and Geiger, Atticus and Geva, Mor and Hoogland, Jesse and Murfet, Daniel and McGrath, Tom},
  year = 2025,
  month = jan,
  number = {arXiv:2501.16496},
  eprint = {2501.16496},
  primaryclass = {cs},
  publisher = {arXiv},
  doi = {10.48550/arXiv.2501.16496},
  urldate = {2025-02-18},
  archiveprefix = {arXiv}
}

@misc{bereskaMechanisticInterpretabilityAI2024,
  title = {Mechanistic {{Interpretability}} for {{AI Safety}} -- {{A Review}}},
  author = {Bereska, Leonard and Gavves, Efstratios},
  year = 2024,
  month = aug,
  number = {arXiv:2404.14082},
  eprint = {2404.14082},
  primaryclass = {cs},
  publisher = {arXiv},
  urldate = {2024-10-20},
  archiveprefix = {arXiv}
}

@misc{raiPracticalReviewMechanistic2025,
  title = {A {{Practical Review}} of {{Mechanistic Interpretability}} for {{Transformer-Based Language Models}}},
  author = {Rai, Daking and Zhou, Yilun and Feng, Shi and Saparov, Abulhair and Yao, Ziyu},
  year = 2025,
  month = oct,
  number = {arXiv:2407.02646},
  eprint = {2407.02646},
  primaryclass = {cs},
  publisher = {arXiv},
  doi = {10.48550/arXiv.2407.02646},
  urldate = {2025-12-28},
  archiveprefix = {arXiv}
}

@misc{liGenerationJudgmentOpportunities2025,
  title = {From {{Generation}} to {{Judgment}}: {{Opportunities}} and {{Challenges}} of {{LLM-as-a-judge}}},
  shorttitle = {From {{Generation}} to {{Judgment}}},
  author = {Li, Dawei and Jiang, Bohan and Huang, Liangjie and Beigi, Alimohammad and Zhao, Chengshuai and Tan, Zhen and Bhattacharjee, Amrita and Jiang, Yuxuan and Chen, Canyu and Wu, Tianhao and Shu, Kai and Cheng, Lu and Liu, Huan},
  year = 2025,
  month = sep,
  number = {arXiv:2411.16594},
  eprint = {2411.16594},
  primaryclass = {cs},
  publisher = {arXiv},
  doi = {10.48550/arXiv.2411.16594},
  urldate = {2026-01-07},
  archiveprefix = {arXiv}
}

@misc{thakurJudgingJudgesEvaluating2025,
  title = {Judging the {{Judges}}: {{Evaluating Alignment}} and {{Vulnerabilities}} in {{LLMs-as-Judges}}},
  shorttitle = {Judging the {{Judges}}},
  author = {Thakur, Aman Singh and Choudhary, Kartik and Ramayapally, Venkat Srinik and Vaidyanathan, Sankaran and Hupkes, Dieuwke},
  year = 2025,
  month = aug,
  number = {arXiv:2406.12624},
  eprint = {2406.12624},
  primaryclass = {cs},
  publisher = {arXiv},
  doi = {10.48550/arXiv.2406.12624},
  urldate = {2026-01-07},
  archiveprefix = {arXiv}
}

@misc{sunHyperDASAutomatingMechanistic2025,
  title = {{{HyperDAS}}: {{Towards Automating Mechanistic Interpretability}} with {{Hypernetworks}}},
  shorttitle = {{{HyperDAS}}},
  author = {Sun, Jiuding and Huang, Jing and Baskaran, Sidharth and D'Oosterlinck, Karel and Potts, Christopher and Sklar, Michael and Geiger, Atticus},
  year = 2025,
  month = apr,
  number = {arXiv:2503.10894},
  eprint = {2503.10894},
  primaryclass = {cs},
  publisher = {arXiv},
  doi = {10.48550/arXiv.2503.10894},
  urldate = {2025-07-14},
  archiveprefix = {arXiv}
}

@misc{wuAxBenchSteeringLLMs2025,
  title = {{{AxBench}}: {{Steering LLMs}}? {{Even Simple Baselines Outperform Sparse Autoencoders}}},
  shorttitle = {{{AxBench}}},
  author = {Wu, Zhengxuan and Arora, Aryaman and Geiger, Atticus and Wang, Zheng and Huang, Jing and Jurafsky, Dan and Manning, Christopher D. and Potts, Christopher},
  year = 2025,
  month = mar,
  number = {arXiv:2501.17148},
  eprint = {2501.17148},
  primaryclass = {cs},
  publisher = {arXiv},
  doi = {10.48550/arXiv.2501.17148},
  urldate = {2025-06-18},
  archiveprefix = {arXiv},
  langid = {english}
}

@article{wang2025logitlens4llms,
  title={Logitlens4llms: Extending logit lens analysis to modern large language models},
  author={Wang, Zhenyu},
  journal={arXiv preprint arXiv:2503.11667},
  year={2025}
}

@misc{niranjanLimitationsSteeringLanguage2025,
  title = {On the {{Limitations}} of {{Steering}} in {{Language Model Alignment}}},
  author = {Niranjan, Chebrolu and Jaidka, Kokil and Yeo, Gerard Christopher},
  year = 2025,
  month = may,
  number = {arXiv:2505.01162},
  eprint = {2505.01162},
  primaryclass = {cs},
  publisher = {arXiv},
  doi = {10.48550/arXiv.2505.01162},
  urldate = {2025-08-25},
  archiveprefix = {arXiv}
}

@misc{haoPatternsMechanismsContrastive2025,
  title = {Patterns and {{Mechanisms}} of {{Contrastive Activation Engineering}}},
  author = {Hao, Yixiong and Panda, Ayush and Shabalin, Stepan and Ali, Sheikh Abdur Raheem},
  year = 2025,
  month = may,
  number = {arXiv:2505.03189},
  eprint = {2505.03189},
  primaryclass = {cs},
  publisher = {arXiv},
  doi = {10.48550/arXiv.2505.03189},
  urldate = {2026-01-07},
  archiveprefix = {arXiv}
}

@misc{joshiIdentifiableSteeringSparse2025,
  title = {Identifiable {{Steering}} via {{Sparse Autoencoding}} of {{Multi-Concept Shifts}}},
  author = {Joshi, Shruti and Dittadi, Andrea and Lachapelle, S{\'e}bastien and Sridhar, Dhanya},
  year = 2025,
  month = feb,
  number = {arXiv:2502.12179},
  eprint = {2502.12179},
  primaryclass = {cs},
  publisher = {arXiv},
  doi = {10.48550/arXiv.2502.12179},
  urldate = {2025-09-03},
  archiveprefix = {arXiv},
  langid = {english}
}

@article{turner2023steering,
  title={Steering language models with activation engineering},
  author={Turner, Alexander Matt and Thiergart, Lisa and Leech, Gavin and Udell, David and Vazquez, Juan J and Mini, Ulisse and MacDiarmid, Monte},
  journal={arXiv preprint arXiv:2308.10248},
  year={2023}
}

@inproceedings{zhang-etal-2025-personalized,
    title = "Personalized Text Generation with Contrastive Activation Steering",
    author = "Zhang, Jinghao  and
      Liu, Yuting  and
      Wang, Wenjie  and
      Liu, Qiang  and
      Wu, Shu  and
      Wang, Liang  and
      Chua, Tat-Seng",
    editor = "Che, Wanxiang  and
      Nabende, Joyce  and
      Shutova, Ekaterina  and
      Pilehvar, Mohammad Taher",
    booktitle = "Proceedings of the 63rd Annual Meeting of the Association for Computational Linguistics (Volume 1: Long Papers)",
    month = jul,
    year = "2025",
    address = "Vienna, Austria",
    publisher = "Association for Computational Linguistics",
    url = "https://aclanthology.org/2025.acl-long.353/",
    doi = "10.18653/v1/2025.acl-long.353",
    pages = "7128--7141",
    ISBN = "979-8-89176-251-0"
}

@inproceedings{rimsky-etal-2024-steering,
    title = "Steering Llama 2 via Contrastive Activation Addition",
    author = "Rimsky, Nina  and
      Gabrieli, Nick  and
      Schulz, Julian  and
      Tong, Meg  and
      Hubinger, Evan  and
      Turner, Alexander",
    editor = "Ku, Lun-Wei  and
      Martins, Andre  and
      Srikumar, Vivek",
    booktitle = "Proceedings of the 62nd Annual Meeting of the Association for Computational Linguistics (Volume 1: Long Papers)",
    month = aug,
    year = "2024",
    address = "Bangkok, Thailand",
    publisher = "Association for Computational Linguistics",
    url = "https://aclanthology.org/2024.acl-long.828/",
    doi = "10.18653/v1/2024.acl-long.828",
    pages = "15504--15522"
}

@misc{karvonenEmergentWorldModels2024,
  title = {Emergent {{World Models}} and {{Latent Variable Estimation}} in {{Chess-Playing Language Models}}},
  author = {Karvonen, Adam},
  year = 2024,
  month = jul,
  number = {arXiv:2403.15498},
  eprint = {2403.15498},
  primaryclass = {cs},
  publisher = {arXiv},
  urldate = {2024-10-21},
  archiveprefix = {arXiv}
}

@misc{tanAnalyzingGeneralizationReliability2025,
  title = {Analyzing the {{Generalization}} and {{Reliability}} of {{Steering Vectors}}},
  author = {Tan, Daniel and Chanin, David and Lynch, Aengus and Kanoulas, Dimitrios and Paige, Brooks and {Garriga-Alonso}, Adria and Kirk, Robert},
  year = 2025,
  month = may,
  number = {arXiv:2407.12404},
  eprint = {2407.12404},
  primaryclass = {cs},
  publisher = {arXiv},
  doi = {10.48550/arXiv.2407.12404},
  urldate = {2025-08-25},
  archiveprefix = {arXiv}
}

@misc{wangEnhancingLLMSteering2025,
  title = {Enhancing {{LLM Steering}} through {{Sparse Autoencoder-Based Vector Refinement}}},
  author = {Wang, Anyi and Wu, Xuansheng and Shu, Dong and Ma, Yunpu and Liu, Ninghao},
  year = 2025,
  month = oct,
  number = {arXiv:2509.23799},
  eprint = {2509.23799},
  primaryclass = {cs},
  publisher = {arXiv},
  doi = {10.48550/arXiv.2509.23799},
  urldate = {2025-10-12},
  archiveprefix = {arXiv}
}

@misc{aradSAEsAreGood2025,
  title = {{{SAEs Are Good}} for {{Steering}} -- {{If You Select}} the {{Right Features}}},
  author = {Arad, Dana and Mueller, Aaron and Belinkov, Yonatan},
  year = 2025,
  month = may,
  number = {arXiv:2505.20063},
  eprint = {2505.20063},
  primaryclass = {cs},
  publisher = {arXiv},
  doi = {10.48550/arXiv.2505.20063},
  urldate = {2025-08-26},
  archiveprefix = {arXiv}
}

@misc{lieberumGemmaScopeOpen2024,
  title = {Gemma {{Scope}}: {{Open Sparse Autoencoders Everywhere All At Once}} on {{Gemma}} 2},
  shorttitle = {Gemma {{Scope}}},
  author = {Lieberum, Tom and Rajamanoharan, Senthooran and Conmy, Arthur and Smith, Lewis and Sonnerat, Nicolas and Varma, Vikrant and Kram{\'a}r, J{\'a}nos and Dragan, Anca and Shah, Rohin and Nanda, Neel},
  year = 2024,
  month = aug,
  number = {arXiv:2408.05147},
  eprint = {2408.05147},
  primaryclass = {cs},
  publisher = {arXiv},
  doi = {10.48550/arXiv.2408.05147},
  urldate = {2025-07-18},
  archiveprefix = {arXiv}
}

@misc{choCorrSteerSteeringImproves2025,
  title = {{{CorrSteer}}: {{Steering Improves Task Performance}} and {{Safety}} in {{LLMs}} through {{Correlation-based Sparse Autoencoder Feature Selection}}},
  shorttitle = {{{CorrSteer}}},
  author = {Cho, Seonglae and Wu, Zekun and Koshiyama, Adriano},
  year = 2025,
  month = aug,
  number = {arXiv:2508.12535},
  eprint = {2508.12535},
  primaryclass = {cs},
  publisher = {arXiv},
  doi = {10.48550/arXiv.2508.12535},
  urldate = {2025-08-26},
  archiveprefix = {arXiv},
  langid = {english}
}

@misc{chalnevImprovingSteeringVectors2024,
  title = {Improving {{Steering Vectors}} by {{Targeting Sparse Autoencoder Features}}},
  author = {Chalnev, Sviatoslav and Siu, Matthew and Conmy, Arthur},
  year = 2024,
  month = nov,
  number = {arXiv:2411.02193},
  eprint = {2411.02193},
  primaryclass = {cs},
  publisher = {arXiv},
  doi = {10.48550/arXiv.2411.02193},
  urldate = {2025-07-23},
  archiveprefix = {arXiv}
}

@misc{wangImprovingLLMReasoning2025,
  title = {Improving {{LLM Reasoning}} through {{Interpretable Role-Playing Steering}}},
  author = {Wang, Anyi and Shu, Dong and Wang, Yifan and Ma, Yunpu and Du, Mengnan},
  year = 2025,
  month = jun,
  number = {arXiv:2506.07335},
  eprint = {2506.07335},
  primaryclass = {cs},
  publisher = {arXiv},
  doi = {10.48550/arXiv.2506.07335},
  urldate = {2025-09-03},
  archiveprefix = {arXiv}
}

@misc{sooInterpretableSteeringLarge2025,
  title = {Interpretable {{Steering}} of {{Large Language Models}} with {{Feature Guided Activation Additions}}},
  author = {Soo, Samuel and Guang, Chen and Teng, Wesley and Balaganesh, Chandrasekaran and Guoxian, Tan and Ming, Yan},
  year = 2025,
  month = apr,
  number = {arXiv:2501.09929},
  eprint = {2501.09929},
  primaryclass = {cs},
  publisher = {arXiv},
  doi = {10.48550/arXiv.2501.09929},
  urldate = {2025-10-12},
  archiveprefix = {arXiv}
}

@article{nguyenMultiAttributeSteeringLanguage,
  title = {Multi-{{Attribute Steering}} of {{Language Models}} via {{Targeted Intervention}}},
  author = {Nguyen, Duy and Prasad, Archiki and {Stengel-Eskin}, Elias and Bansal, Mohit},
  langid = {english}
}
